\documentclass[lettersize,journal]{IEEEtran}
\usepackage{amsmath,amsfonts}
\usepackage{algorithmic}
\usepackage{algorithm}
\usepackage{array}
\usepackage[caption=false,font=normalsize,labelfont=sf,textfont=sf]{subfig}
\usepackage{textcomp}
\usepackage{stfloats}
\usepackage{url}
\usepackage{verbatim}
\usepackage{graphicx}
\usepackage{cite}
\usepackage{xcolor}
\usepackage{multirow}
\usepackage{adjustbox}
\usepackage{mathrsfs}
\usepackage{booktabs}
\usepackage{hyperref}
\usepackage{epstopdf}
\begin{document}

\title{UGD: An Unsupervised Geometric Distance for Evaluating Real-world Noisy Point Cloud Denoising}

\author{Zhiyong~Su,
	    Jincan Wu,
        Yanke Li,
	    Yonghui Liu,
	    Zheng Li,
	    and~Weiqing~Li % <-this % stops a space
\IEEEcompsocitemizethanks{\IEEEcompsocthanksitem Z. Su, J. Wu, Y. Li, Y. Liu, and Z. Li are with School of Automation, Nanjing University of Science and Technology, Nanjing 210094, China.\protect\\
E-mails: su@njust.edu.cn, 2500381260@qq.com, 1327975414@qq.com, 2146623762@qq.com, 1315571938@qq.com
\IEEEcompsocthanksitem W. Li is with School of Computer Science and Engineering, Nanjing University of Science and Technology, Nanjing 210094, China.\protect\\
E-mail: li\_weiqing@njust.edu.cn}% <-this % stops an unwanted space
\thanks{Manuscript received April 19, 2005; revised August 26, 2015. (Corresponding author: Weiqing Li.)}}

\markboth{Journal of \LaTeX\ Class Files,~Vol.~14, No.~8, August~2021}%
{Shell \MakeLowercase{\textit{et al.}}: A Sample Article Using IEEEtran.cls for IEEE Journals}


\maketitle

\begin{abstract}
Point cloud denoising is a fundamental and crucial challenge in real-world point cloud applications.
Existing quantitative evaluation metrics for point cloud denoising methods are implemented in a supervised manner, which requires both the denoised point cloud and the corresponding ground-truth clean point cloud to compute a representative geometric distance.
This requirement is highly problematic in real‑world scenarios, where ground‑truth clean point clouds are often unavailable.
In this paper, we propose a simple yet effective unsupervised geometric distance (UGD) for real-world noisy point cloud denoising, calculated solely from noisy point clouds.
The core idea of UGD is to learn a patch-wise prior model from a set of clean point clouds and then employ this prior model as the ground-truth to quantify the degradation by measuring the geometric variations of the denoised point cloud.
To this end, we first learn a pristine Gaussian Mixture Model (GMM) with extracted patch-wise quality-aware features from a set of pristine clean point clouds by a patch-wise feature extraction network, which serves as the ground-truth for the quantitative evaluation.
Then, the UGD is defined as the weighted sum of distances between each patch of the denoised point cloud and the learned pristine GMM model in the patch space.
To train the employed patch-wise feature extraction network, we propose a self-supervised training framework through multi-task learning, which includes pair-wise quality ranking, distortion classification, and distortion distribution prediction.
Quantitative experiments with synthetic noise confirm that the proposed UGD achieves comparable performance to supervised full-reference metrics. 
Moreover, experimental results on real-world data demonstrate that the proposed UGD enables unsupervised evaluation of point cloud denoising methods based exclusively on noisy point clouds.
The source code is released at \href{https://github.com/Takahashi314/UGD}{https://github.com/Takahashi314/UGD}.
\end{abstract}


\section{Introduction}
\label{sec:intro}

Point cloud denoising is a fundamental challenge in the field of computer graphics. 
As a key preprocessing step, it plays a crucial role in downstream tasks, e.g., registration and reconstruction, since various types of noise may be introduced during the acquisition, compression, and transmission processes.
To tackle this problem, recently, numerous supervised point cloud denoising methods have been proposed, which are trained with synthetic datasets consisting of clean point clouds and their corrupted versions with simulated noises \cite{edirimuni-2024-contrastive,wei2024pathnet}.
Besides, due to the lack of ground-truth clean point clouds in real-world scenarios, a few unsupervised point cloud denoising algorithms have also been developed \cite{hermosilla2019total,su2025cad}.
Existing supervised and unsupervised point cloud denoising approaches have demonstrated impressive results on both synthetic and real-world noisy point clouds \cite{Zhoul22}.
Therefore, how to objectively and comprehensively benchmark point cloud denoising methods remains a fundamental and challenging task.

However, to the best of our knowledge, no unsupervised geometric metrics currently exist to quantitatively evaluate the geometric quality of denoised point clouds using only noisy point clouds.
Quantitative evaluation of point cloud denoising methods is currently dominated by supervised geometric metrics, including Hausdorff Distance (HD) \cite{javaheri2020generalized, lavoue2010comparison}, Mean Squared Error (MSE) \cite{lavoue2010comparison}, Peak Signal-to-Noise Ratio (PSNR) \cite{tian2017geometric, garg2019survey}, Chamfer Distance (CD) \cite{zhang2020pointfilter}, etc.
These geometric evaluation metrics typically quantify the geometric error in a supervised fashion. 
Specifically, they calculate the geometric distance between denoised point clouds and their corresponding ground-truth clean point clouds.
However, in real-world scenarios, the ground-truth clean point clouds required by these supervised geometric metrics are often unavailable. Consequently, the evaluation can only be carried out in an unsupervised way, i.e., exclusively from the noisy point clouds.
As a result, only qualitative evaluation can be performed to evaluate the denoising performance on real-world noisy point clouds through subjective visual inspection, lacking the ability to provide quantitative analytical results \cite{marcos2023evaluating,Zhoul22}.
Therefore, the absence of unsupervised geometric metrics significantly hinders the development and practical application of real-world noisy point cloud denoising approaches.

Currently, no-reference (NR) point cloud quality assessment (PCQA) methods have been widely studied for evaluating the perceptual quality of distorted colored point clouds, which do not need the assistance of reference ground-truth samples \cite{zyj23, liu2021pqa, shan2024contrastive}.
These approaches, which are trained on subjective datasets with Mean Opinion Score (MOS), target to predict visual quality scores (e.g., 1-5) for input colored point clouds.
To this end, both color and geometric distortions must be considered in constructing subjective datasets and predicting visual quality scores.
However, existing point cloud denoising approaches are primarily designed for colorless point clouds \cite{ feng2024forest, wei2024pathnet}.
Color distortions, as well as other geometric degradation(e.g., downsampling and compression)  considered in NR PCQA area, are not included in point cloud denoising.
Most importantly, rather than assigning perceptual quality scores, this paper aims to quantify the geometric distance between the underlying ground-truth and the observed noisy measurements (i.e., point clouds).
Therefore, existing NR PCQA methods are unsuitable for benchmarking point cloud denoising approaches. 

To address the above issues, this paper proposes a novel unsupervised geometric distance (UGD) to evaluate real-world noisy point cloud denoising, which is computed exclusively from input noisy point clouds.
The core idea is to learn a prior model that serves as the ground-truth from a set of clean point clouds, and then quantify the degradation through measuring the geometric quality variations of the denoised point cloud from the learned priors.
Specifically, we first collect a set of high-quality clean point clouds and extract patch-wise geometric quality-aware features using a feature extraction network.
Then, we learn a Gaussian Mixture Model (GMM) from extracted patch-wise features as the prior model, which serves as the ground-truth.
Finally, given a denoised point cloud, the UGD is defined as the weighted sum of distances between each patch and the learned pristine GMM model.
Moreover, to train the feature extraction network, we also introduce a self-supervised multi-task training framework, which utilizes pair-wise ranking, distortion classification, and distortion distribution prediction tasks to guide the network to learn geometric quality-aware features.
The key contributions of this paper are as follows:
\begin{itemize}
    \item 
    We propose a novel unsupervised geometric distance to evaluate real-world noisy point cloud denoising in the absence of ground-truth clean point clouds.
    The UGD is computed by weighting the sum of distances between each patch of the denoised point cloud and the learned pristine prior GMM, using only the denoised point cloud.
    \item 
    We present a self-supervised multi-task training framework for the geometric quality-aware feature extraction network to learn a patch-wise prior, which integrates pair-wise ranking, distortion classification, and distortion distribution prediction tasks. 
    % \item    
    % Experimental results demonstrate that the proposed UGD enables unsupervised evaluation of denoising methods based exclusively on denoised point clouds.
    % Additionally, the proposed UGD performs comparably to existing supervised evaluation metrics on synthetic datasets.
\end{itemize}

\section{Related Works} 
\label{sec:related_works}

\subsection{Point Cloud Denoising}

Recently, numerous works on point cloud denoising have been presented and have achieved promising performance \cite{Zhoul22,Wang24,Edirimuni2024denoise,Edirimuni2023IterativePFN}.
These approaches can be categorized into three groups: traditional point cloud denoising, supervised point cloud denoising, and unsupervised point cloud denoising.

Traditional point cloud denoising methods primarily focus on Moving Least Squares (MLS) approaches \cite{alexa-2003-MLS,fleishman-2005-RMLS, oztireli-2009-RIMLS} and Locally Optimal Projection (LOP) methods \cite{lipman-2007-LOP,huang-2009-WLOP,preiner-2014-CLOP}.
These methods  typically involve numerous parameters and require meticulous trial-and-error parameter tuning to achieve satisfactory denoising results, as the geometric structures of the point cloud are complex and the noise characteristics are often unknown.

Supervised point cloud denoising approaches aim to learn a mapping from the noisy inputs to their ground-truth clean counterparts and have achieved notable success.
These methods include pointnet-based \cite{guerrero-2018-pcpnet,rakotosaona2020pointcleannet,Huang2023MODNet}, convolution-based \cite{wei-2021-geodualcnn}, encoder-decoder-based methods \cite{zhang2020pointfilter}.
Furthermore, generative models have emerged as a promising direction. For instance, score-based approaches \cite{Luo2021Score-based}, normalizing flow-based frameworks \cite{unknownue2022pdflow}, and invertible neural networks in latent space \cite{Mao2024LatentSpace} have been proposed to better model the underlying distribution of clean surfaces.
However, these approaches generally require extensive training on large-scale synthetic datasets, which are typically generated by artificially introducing various noise types into clean point clouds. 
As a result, the performance of these methods may deteriorate significantly when applied to point clouds that deviate markedly from the training data, such as real-world noisy point clouds.

Unsupervised point cloud denoising methods aim to directly remove noise from point clouds without relying on paired clean and noisy point clouds. 
Unlike supervised approaches, which typically require large datasets of paired clean and noisy examples, unsupervised methods address the challenge of limited ground-truth data. 
To date, only a limited number of unsupervised denoising methods have been proposed for point clouds \cite{hermosilla2019total,luo-2020-DMRDenoise,su2025cad,Wang2024Noise4Denoise}.

Overall, existing point cloud denoising methods have exhibited remarkable performance on both synthetic and real-world datasets. 
However, no geometric metrics exist to evaluate these methods in an unsupervised fashion. 
This deficiency poses a critical challenge for numerous practical applications, as ground-truth clean point clouds are frequently unavailable. 
In such cases, visual inspection is the only available means of assessment, which is both subjective and insufficient for rigorous evaluation.
This paper is dedicated to designing a novel unsupervised geometric metric, computed exclusively from noisy data, to evaluate real-world noisy point cloud denoising.

\subsection{Existing Supervised Evaluation Metrics}

Existing supervised evaluation metrics calculate the distance between noisy point clouds and their corresponding ground-truth clean point clouds to quantitatively assess geometric errors.
These metrics can be categorized into two main types: distance-based and normal-based metrics.
Distance-based metrics can be further divided into three categories: point-to-point, point-to-plane, and point-to-mesh.
Normal-based metrics are also referred to as plane-to-plane.

Point-to-point approaches (po2po) \cite{Rufaelm16} assess the distance between points in the noisy point cloud and their corresponding points in the reference clean point cloud. 
Point-to-plane methods (po2pl) \cite{tian2017geometric} are derived from the projection of the vector connecting two associated points along the normal vector of the reference point, and larger costs are assigned to points that deviate from the underlying surface.
Plane-to-plane techniques (pl2pl) \cite{Alexioue18icme} are based on the angular similarity of tangent planes that correspond to associated points between the noisy and reference point clouds.
For each metric discussed above, an individual distance is associated with every point of the point cloud under evaluation.
Additionally, the overall geometric similarity between the test point clouds is expressed through a total error value. 
The total error value can be computed using various methods, such as Hausdorff Distance (HD) \cite{lavoue2010comparison}, Root Mean Square Error (RMSE) \cite{javaheri2017subjective}, Peak Signal-to-Noise Ratio (PSNR) \cite{garg2019survey, tian2017geometric}, and Chamfer Distance (CD) \cite{rakotosaona2020pointcleannet, zhang2020pointfilter, hermosilla2019total}.
Point-to-mesh metrics (po2me) \cite{Cignoni1998Metro} first reconstruct the surface and then measure the projected distances between the evaluated point cloud and the reconstructed reference object. 
However, the efficiency of po2me is heavily dependent on the selected surface reconstruction algorithm.

In summary, current supervised geometric evaluation metrics have demonstrated impressive performance on benchmarks based on synthetic noise.
However, no metrics exist to evaluate point cloud denoising approaches in an unsupervised fashion, which is highly problematic in real-world point cloud applications where ground-truth clean point clouds are often either impossible or very constraining.
Consequently, in the absence of unsupervised metrics, only subjective visual inspections can be employed to evaluate the performance on real-world noisy point clouds. 

\subsection{Point Cloud Quality Assessment}

PCQA methods aim to predict perceptual quality scores of degraded point clouds based on human visual perception \cite{wuxj21}.
PCQA methods can be categorized into three groups according to the availability of reference information: full-reference (FR), reduced-reference (RR), and no-reference (NR).

\subsubsection{FR PCQA methods} 

FR PCQA methods typically involve evaluating the quality of a point cloud by comparing it to its corresponding high-quality reference point cloud \cite{Alexioue18,Yangq22,liuq23,Zhang24TCDM,Zhang24hybrid}.
These approaches offer several advantages, such as accurate evaluation results and strong interpretability, as they can directly reflect the geometric and attribute distortions of the point cloud. 
However, the need for high-quality reference point clouds significantly limits the applicable scenarios of these methods.

\subsubsection{RR PCQA methods} 

RR PCQA methods evaluate the quality of point clouds with limited reference information \cite{viola2020reduced,zhou2023reduced}.  
Specifically, these methods begin by extracting partial features from both the reference point cloud and the distorted point cloud. 
Subsequently, they quantify the quality of the point cloud by comparing these extracted features.
In contrast to FR PCQA methods, RR PCQA methods have a reduced reliance on features. 
This characteristic enables them to offer effective quality evaluation even when only partial information from the reference point cloud is available.
However, RR PCQA methods still require partial information from the original point cloud, and are not applicable to situations where reference point clouds cannot be obtained.

\subsubsection{NR PCQA methods} 

NR PCQA approaches evaluate the quality of the target point cloud without requiring any reference or original point cloud for comparison.
These methods can be subdivided into projection-based and point-based methods.

Projection-based methods evaluate the quality of point clouds by projecting them onto 2D planes and leveraging established 2D image quality assessment metrics. 
PQA-Net \cite{liu2021pqa} employs a multi-view-based joint feature extraction and fusion (MVFEF) module and a distortion type identification (DTI) module to extract distortion features from regular grid images generated by projecting point clouds onto six views. 
IT-PCQA \cite{yang2022no} fuses the images of six views into a single image and utilizes unsupervised adversarial learning to learn prior knowledge from an image quality assessment database, thereby guiding point cloud quality evaluation. 
Zhang et al. \cite{zhang24gms} proposed the NR method GMS-3DQA, which employs the Grid Mini-patch Sampling (GMS) strategy and uses the lightweight Swin-Transformer tiny as the feature extraction backbone. 

Point-based approaches directly process the raw point cloud to evaluate its quality.
These methods focus on extracting quality-aware features from the geometric and color information of point clouds.
Inspired by the hierarchical organization of perception systems and guided by the intrinsic properties of point clouds, ResSCNN \cite{Liu2023} develops a sparse convolutional neural network-based NR-PCQA model. 
GPA-Net \cite{shan24gpanet} employs a novel graph convolution kernel that effectively captures both structural and textural perturbations.
3DTA \cite{zhu243dta} introduces a two-stage sampling method that can reasonably represent an entire point cloud, making it possible to calculate the point cloud quality efficiently.

In summary, current PCQA methods primarily focus on establishing a mapping between the human visual system (HVS) and the perceptual quality of point clouds. 
However, they do not address whether this mapping is consistent with an underlying ground-truth corresponding to the observed noisy measurements (i.e., point clouds). 
This consistency is the primary objective of the proposed metric in this study.

\section{Method} 
\label{sec:method}

\begin{figure*}[!ht]
    \centering
    \includegraphics[width=\textwidth]
    {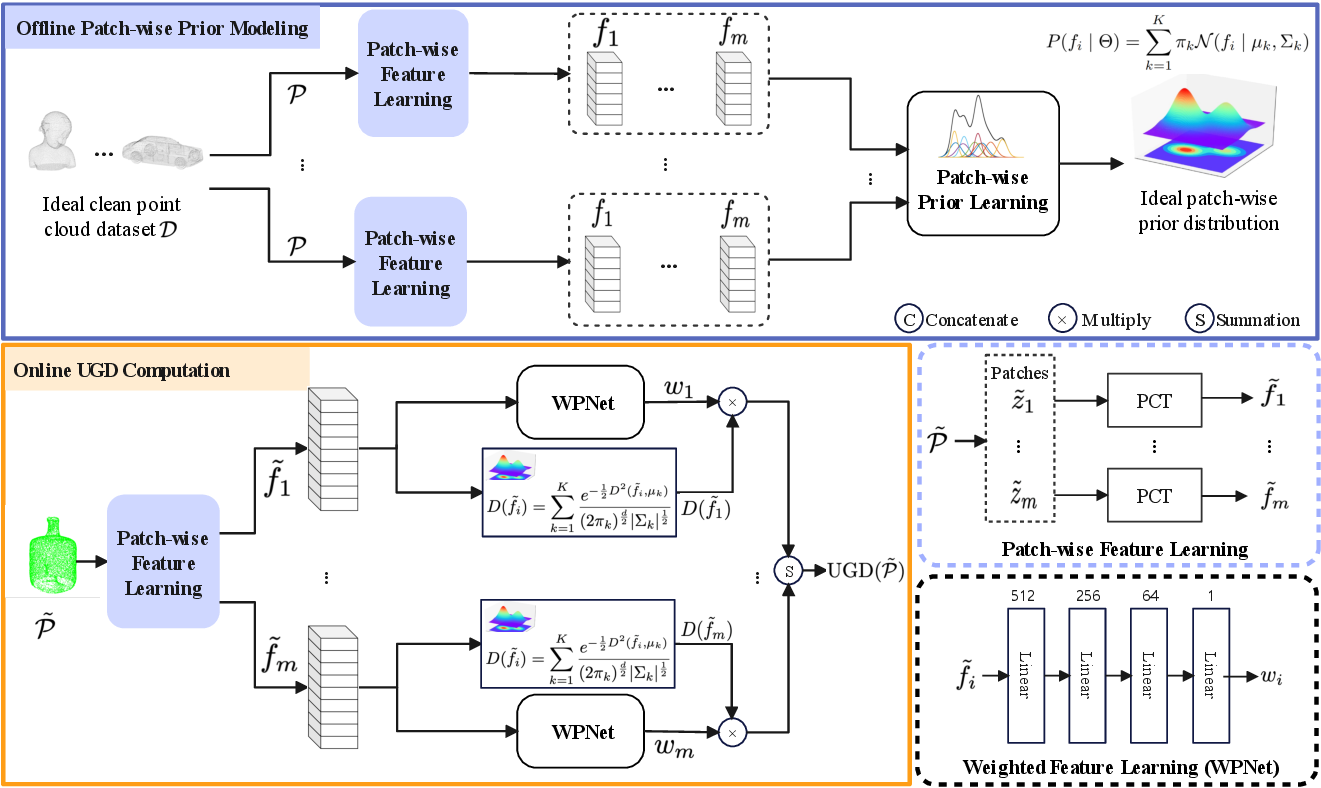}
    \caption{Overview of the proposed unsupervised geometric distance (UGD) approach. It comprises Offline Patch-wise Prior Modeling (top), which learns a GMM prior from clean patch features, and Online UGD Computation (bottom), which quantifies degradation by calculating the weighted Mahalanobis distance between patches of denoised point cloud and the learned prior.}
    \label{fig:overview_of_NRMD}
\end{figure*}

\subsection{Overview}
\label{subsec:Overview}

The proposed UGD metric targets to quantify the degradation of input denoised point cloud through measuring the geometric variations from a learned pristine prior reference, as illustrated in Fig. \ref{fig:overview_of_NRMD}.
To this end, firstly, the offline patch-wise prior modeling is designed to learn a Gaussian Mixture Model (GMM) over patches from high-quality clean point clouds as the prior model, which serves as the ground-truth in the offline phase.
Then, in the online phase, given an input denoised point cloud, the UGD is computed by weighting the sum of distances between each patch of the denoised point cloud and the learned pristine prior GMM model, using only the denoised point cloud.

\subsection{Offline Patch-wise Prior Modeling}

To model the underlying patch distribution of clean point clouds in the offline phase, firstly, the patch-wise feature learning extracts informative patch-wise geometric quality-aware features from a set of ideal clean point clouds $\mathcal{D}$.
Then, the patch-wise prior learning learns a pristine GMM model from the extracted patch-wise prior knowledge to serve as a reference in the latent space for further quality calculation.

\subsubsection{Ideal Clean Point Cloud Dataset}

\begin{figure}[!t]
	\centering
	\includegraphics[width=0.45\textwidth]{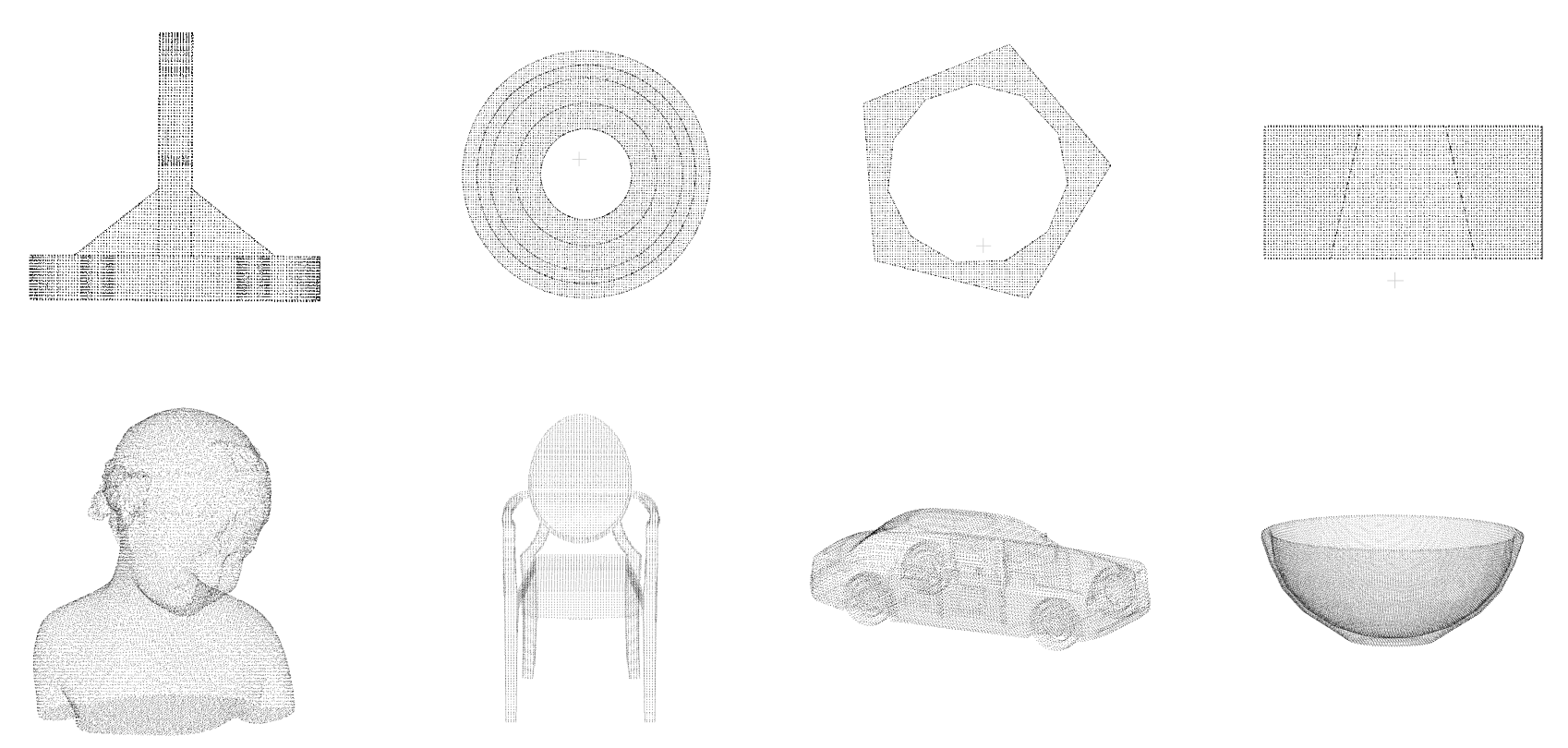}    
	\caption{Snapshots of some point clouds in the ideal clean point cloud dataset $\mathcal{D}$.}
	\label{fig:dataset_show}
\end{figure}

The ideal clean point cloud dataset $\mathcal{D}$ consists of diverse, high-quality pristine point clouds.
Specifically, the proposed dataset $\mathcal{D}$ contains 150 high-quality clean reference point clouds with diverse geometric complexity, including 100 regular-shaped industrial parts and 50 irregular-shaped objects, covering various categories such as vehicles, people, and daily necessities, as illustrated in Fig. \ref{fig:dataset_show}.
All selected point clouds come from existing 3D mesh datasets, including Stanford 3D Scanning Repository \cite{TurkG94} and ModelNet \cite{Wuzr15}.
For each selected 3D mesh, a uniformly distributed random sampling strategy is employed to obtain the Cartesian coordinates of the generated point cloud from the mesh surfaces. 

\subsubsection{Patch-wise Feature Learning}
\label{subsubsec:Patch-Wise Feature Learning}

The patch-wise feature learning is designed to partition input point clouds into overlapped patches, and then learn patch-wise geometric quality-aware features for each patch through a well-trained feature extraction network.
%The extracted geometric quality-aware features from clean point clouds are used to establish an ideal model as the clean ground-truth prior.

To generate patches from each clean point cloud $\mathcal{P}$, firstly, the farthest point sampling (FPS) \cite{qi2017pointnet++} technique is used to select $m$ anchor points $p_{i}^a, i \in [1,m]$. 
After that, for each anchor point $p_{i}^a$, $s$ points are randomly selected within a sphere of radius $r$ centered at $p_{i}^a$ to form a patch $z_i$.
Those patches whose point numbers are less than $s$ are padded to $s$ points by randomly selecting points from their existing points inside the referred sphere.

To extract geometric quality-aware features $f_i$ of each path $z_i$, the widely used Point Cloud Transformer (PCT) \cite{guo2021pct} is employed in this paper.
However, the original PCT is trained to specialize towards solving point cloud classification and segmentation tasks, which are fundamentally different from the geometric quality evaluation task discussed in this paper.
For instance, the feature representations learned by the original PCT are expected to be distortion-invariant for the classification and segmentation tasks.
Conversely, for the geometric quality evaluation task, the extracted features should be sensitive to various degradations, thereby facilitating the evaluation of geometric quality of denoised point clouds.
Therefore, a self-supervised multi-task training strategy is introduced to train the employed feature extraction network, which will be discussed in Section \ref{sec:ss}.

\subsubsection{Patch-wise Prior Learning}
\label{subsec:Patch Distribution Learning}

The patch-wise prior modeling aims to learn a finite GMM over patches from the ideal clean point cloud dataset $\mathcal{D}$ to capture the underlying distribution of clean geometric structures. 
The learned prior serves as a reference for subsequent online UGD computation.

First, as discussed in Section \ref{subsubsec:Patch-Wise Feature Learning}, we can get $m$ overlapped patches from each clean point cloud $\mathcal{P}$ in the dataset $\mathcal{D}$.
Thus, a total of $S = 150 \times m$ patches can be extracted from the dataset $\mathcal{D}$, and the patch group can be denoted as $Z = \{z_{i}\}_{i=1}^{S}$.

Then, a finite GMM prior $P(\cdot \mid \Theta)$ is learned to describe the distribution of patch features extracted from the patch group $Z$.
For each patch $z_i$ with its $d$-dimensional feature vector $f_i$, the probability density under the GMM is defined as:
\begin{equation}
\label{eq:e1}
  P(f_i \mid \Theta) = \sum_{k=1}^{K} \pi_k \mathcal{N}(f_i \mid \mu_k, \Sigma_k),
\end{equation}
where $K$ denotes the total number of Gaussian components, and the GMM model is parameterized by
$\Theta = \{\pi_k, \mu_k, \Sigma_k\}_{k=1}^{K}$, representing the mixing coefficients, mean vectors, and covariance matrices, respectively, subject to $\sum_{k=1}^{K}\pi_k=1$.
$\mathcal{N}(\cdot)$ is the Gaussian distribution calculated as
\begin{equation}
\label{eq:gaussian}
    \mathcal{N}(f_i \mid \mu_k, \Sigma_k) =
    \frac{\mathrm{exp}\left(-\frac{1}{2}(f_i-\mu_k)^T \Sigma_k^{-1} (f_i-\mu_k)\right)}
    {(2\pi)^{\frac{d}{2}}\left | \Sigma_k  \right | ^{\frac{1}{2}} }.
\end{equation}
Assuming that all patches in the group $Z$ are independent, the joint likelihood function $\mathcal{L}$ of the patch features $\{f_i\}_{i=1}^{S}$ under the GMM can be written as:
\begin{equation}
\label{eq:e3}
    \mathcal{L}(\Theta) = \prod_{i=1}^{S} P(f_i \mid \Theta)
    = \prod_{i=1}^{S} \left( \sum_{k=1}^{K} \pi_k \mathcal{N}(f_i \mid \mu_k, \Sigma_k) \right).
\end{equation}
Thus, the parameters $\Theta$ can be determined by maximum likelihood, typically using the Expectation Maximization algorithm \cite{dempster1977maximum}.

Finally, the patch-wise prior of clean point clouds is modeled by the learned GMM $P(\cdot \mid \Theta)$.
In the subsequent UGD computation, each patch feature (from a noisy or denoised point cloud) can be evaluated by its likelihood under this learned prior, and each Gaussian component $\{\mu_k, \Sigma_k\}$ provides a local statistical description in the patch-feature space.

\subsection{Online UGD Computation}
\label{subsec:UGD Computation}

%2020-SP-Unsupervised quaternion model for blind colour image quality assessment.pdf
%3.1. Framework of the proposed method
Given an input denoised point cloud $ \tilde{\mathcal{P}} $, its UGD is defined as the weighted sum of the Mahalanobis-distance-based scores between each patch and the learned GMM $P(\cdot \mid \Theta)$ in the patch space.
%computed by measuring the likelihood of patch-wise features falling at the center of the prior Gaussian distribution.

Firstly, we divide the input denoised point cloud $\tilde{\mathcal{P}}$ into $m$ overlapped patches $\tilde{z}_i \  (i \in [1,m])$, and extract patch-wise features $ \tilde{f}_i $ of $\tilde{z}_i$, as discussed in Section \ref{subsubsec:Patch-Wise Feature Learning}.
Then, the Mahalanobis distance $ D (\tilde{f}_i, \mu_k) $ between the patch $\tilde{z}_i$ and the center of the $k$-th Gaussian component $\mu_k$ can be calculated by:
\begin{equation}
D (\tilde{f}_i, \mu_k) = \sqrt{(\tilde{f}_i-\mu_k)^T\Sigma_k^{-1}(\tilde{f}_i-\mu_k)}.
\end{equation}
Subsequently, the Mahalanobis-distance-based score $ D (\tilde{f}_i) $ between each patch $\tilde{z}_i$ and the learned GMM $P(\cdot \mid \Theta)$ can be obtained by:
\begin{equation}
D (\tilde{f}_i) = \sum_{k=1}^{K} \pi_k
\frac{\exp\left(- \frac{1}{2} D^2(\tilde{f}_i, \mu_k)\right)}{(2\pi)^{\frac{d}{2} }| \Sigma_k |^{\frac{1}{2}}},
\end{equation}
where $\pi_k$ is the mixing coefficient of the $k$-th Gaussian component, and $d$ is the dimension of $\tilde{f}_i$.
Finally, the UGD of the point cloud $ \tilde{\mathcal{P}} $ is defined as the weighted Mahalanobis-distance-based score between each patch and the learned GMM $P(\cdot \mid \Theta)$:
\begin{equation}
\label{eq:ugd}
\mathrm{UGD} ( \tilde{\mathcal{P}}) = \sum_{i=1}^{m} w_i \cdot D (\tilde{f}_i),
\end{equation}
where $w_{i} = \mathrm{WPNet}(\tilde{f}_i)$ is the weight that represents the contribution of the patch $\tilde{f}_i$, which is defined in Section \ref{subsec:Feature Extraction}.

It is noted that UGD is a relative distance metric, which indicates that it can only be meaningful when compared within the same point cloud.
The higher the UGD, the higher the quality.

\begin{figure*}[!ht]
	\centering
	\includegraphics[width=1\textwidth]{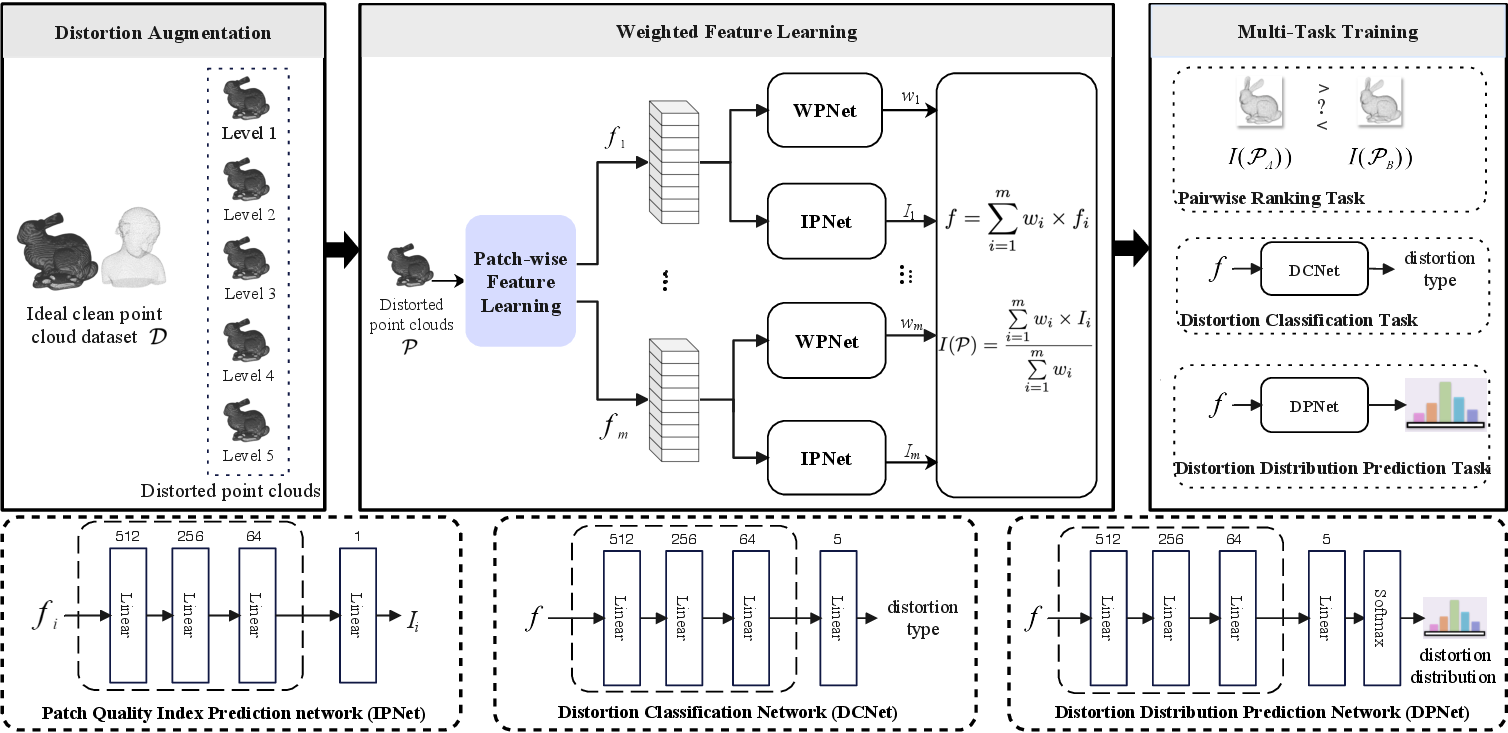}
	\caption{Overview of the Self-Supervised Multi-Task (SSMT) training framework. It integrates Distortion Augmentation (generating diverse noise samples), Weighted Feature Learning (extracting adaptive patch features), and Multi-Task Training (Ranking, Classification, and Prediction tasks) to learn robust geometric quality-aware representations.}
	\label{fig:overview_of_SSTM}
\end{figure*}

\section{Self-supervised Multi-task Training}
\label{sec:ss}

%2022-TIP-Contrastive Self-Supervised Pre-Training for Video Quality Assessment.pdf III. PROPOSED APPROACH
% 参考 2022_SPIQ_A Self-Supervised Pre-Trained Model for Image Quality Assessment
The self-supervised multi-task training (SSMT) aims to train the employed feature encoder in the patch-wise feature learning to extract effective geometric quality-aware representations.
As discussed in Section \ref{subsubsec:Patch-Wise Feature Learning}, the employed feature encoder PCT is pre-trained to specialize towards solving point cloud classification and segmentation tasks, which is significantly distinguished from the geometric quality assessment task.
Besides, existing large scale PCQA datasets with MOS, such as SJTU-PCQA \cite{Yangq21}, SIAT-PCQD \cite{wuxj21}, WPC \cite{liuq23}\cite{su2019perceptual}, WPC2.0  \cite{liu2021reduced}, WPC3.0 \cite{liu2022no}, and LS-PCQA \cite{Liu2023}, involve both geometry and color distortions.
These datasets are tailored for training learning-based PCQA methods to predict visual quality scores of colored point clouds.
Therefore, it is difficult to employ existing subjective PCQA datasets to train the feature encoder for the geometric quality assessment task discussed in this paper.

Fig. \ref{fig:overview_of_SSTM} depicts the overview of the proposed SSMT strategy.
Given each high-quality clean point cloud in the proposed dataset $\mathcal{D}$, we first generate distorted counterparts through the distortion augmentation strategy. 
Then, a weighted feature learning module is designed to extract geometric quality-aware features of the input point cloud.
Finally, the pair-wise ranking task, distortion classification task, and distortion distribution prediction task are introduced to guide the employed network to learn geometric quality-aware feature representations.
The intuition behind this work is that the employed feature encoder PCT cannot effectively perform the three tasks unless it is capable of capturing the geometric distortion-related information. 

\subsection{Distortion Augmentation}
\label{subsec:distortion_augmentation}

% 2022_TIP_Contrastive Self-Supervised Pre-Training for Video Quality Assessment  B. Distortion Augmentation
% 2022_SPIQ_A Self-Supervised Pre-Trained Model for Image Quality Assessment A. Distortion Augmentation
The distortion augmentation is designed to derive an efficient self-supervision signal for conducting the proposed self-supervised training, in which the distorted samples with diverse distortion categories and levels are generated from the clean point clouds in the proposed dataset $\mathcal{D}$.

Specifically, considering the types of noise often considered by existing denoising algorithms, each reference point cloud in the proposed dataset $\mathcal{D}$ is distorted by five different distortions under five levels, including Gaussian noise (GN), uniform noise (UN), impulse noise (IN), exponential noise (EN), and mixed noise (MN).
To this end, for each reference point cloud, the edge length between each point and its nearest neighbor is first calculated. 
Then, the average length $l_r$  of all edges is employed as the reference value to generate distortions.
\begin{itemize}
    \item 
    Gaussian noise (GN): Gaussian distributions with means of $0$ with standard deviations $\xi \times l_r $ ($\xi \in \lbrace 0.2, 0.4, 0.7, 1.0, 1.4 \rbrace$) are added to the reference point cloud.
    \item 
    Uniform noise (UN): Uniform distributions with intervals $\pm (\xi \times l_r) $ ($\xi \in \lbrace 0.6, 1.2, 2.1, 3.0, 4.2 \rbrace$) are added to the reference point cloud.
    \item 
    Impulse noise (IN): Impulse noise is generated by uniform distributions with intervals $\pm (\xi \times l_r) $ ($\xi \in \lbrace 0.6, 1.2, 2.1, 3.0, 4.2 \rbrace$) as the initial distribution. Then, a threshold of $\pm 0.2 l_r$ is set, and values exceeding this threshold will be added to the reference point cloud.
    \item 
    Exponential noise (EN): Exponential distributions with means $\xi \times l_r $ ($\xi \in \lbrace 0.2, 0.4, 0.7, 1.0, 1.4 \rbrace$) are added to the reference point cloud.
    % \item 
    % Simulated real noise (SN): Simulated real noise comes from the time-of-flight camera in the BlenSor software \cite{BlenSor}. Different sampling levels are simulated by different opening angle of the camera among $25^{\circ}$, $35^{\circ}$, $45^{\circ}$, $55^{\circ}$, and $65^{\circ}$, respectively. The smaller the opening angle, the lower the distortion of obtained point clouds.
    \item 
    Mixed noise (MN): Mixed noise is formed by the combination of the above four basic types of noise: $e_{MN} = \sum_{i \in \{GN, UN, IN, EN\}} \alpha_i \times e_i$, where $e_i$ represents the basic noise at a given noise level, $\alpha_i$ denotes the weight of each basic noise in the mixed noise. 
    The weight $\alpha_i$ is initially generated as random value uniformly distributed in the interval $[0,1]$, and then normalized as $\alpha_i = \frac{\alpha_i}{\sqrt{ \alpha_{GN}^2 + \alpha_{UN}^2 + \alpha_{IN}^2 + \alpha_{EN}^2}}$.
\end{itemize}

using five noise types at five levels

\subsection{Weighted Feature Learning}
\label{subsec:Feature Extraction}

The weighted feature learning aims to learn geometric quality-aware features of the input point cloud by weighting the features of all patches extracted through the patch-wise feature learning module discussed in Section \ref{subsubsec:Patch-Wise Feature Learning}.
The intuition behind the weighted feature learning is that not every patch in a point cloud contributes equally to the geometric quality of the entire model.

Given an input point cloud $ \mathcal{P} $, the feature vector $f_i \  (i \in [1,m])$  of each patch $z_i$ of $ \mathcal{P} $ is first extracted through the patch-wise feature learning module as discussed in Section \ref{subsubsec:Patch-Wise Feature Learning}.
Then, a patch quality weight prediction network (WPNet) is introduced to assign an adaptive weight to each patch. 
The WPNet adopts an MLP model, in which the fully connected layer sizes are 512, 256, 64 and 1, respectively.
The Sigmoid layer is employed to output the weight $w_i$ of each patch $z_i$.
Finally, the feature vector $f$ of $ \mathcal{P} $ composed of $m$ patches can be obtained by:
\begin{equation}
\label{eq:fp}
	f = \sum\limits_{i = 1}^m {w_i \times f_i},
\end{equation}
where $w_{i} = \mathrm{WPNet}(f_i)$.

\subsection{Pair-wise Ranking Task}

%2022-SPIC-Twice Mixing-A rank learning based quality assessment approach for underwater image enhancement.pdf

The pair-wise ranking task is designed to learn a ranking function $\mathscr{R}$ to predict the probability that $\tilde{\mathcal{P}}_A$ is better than $\tilde{\mathcal{P}}_B$, assuming that $\tilde{\mathcal{P}}_A$ and $\tilde{\mathcal{P}}_B$ are the two point clouds of the same clean point cloud $\mathcal{P}$ degraded by the same distortion under different levels. 
Therefore, for each distortion type, $C_5^2 = 10$ possible combinations of five noise levels are constructed for each clean point cloud $\mathcal{P}$ for training.

Firstly, given an input point cloud $\mathcal{P}$, a patch quality index prediction network (IPNet) is designed to compute the quality index $I_i$ of each patch $z_i \ (i \in [1,m])$, which shares the same network structure with WPNet.
The overall quality index $I(\mathcal{P})$ of $\mathcal{P}$ can be obtained by weighting the quality index $I_i$ of each patch as follows:
\begin{equation}
\label{eq:sp}
	I(\mathcal{P}) = \frac{{\sum\limits_{i = 1}^m {w_i \times I_i} }}{{\sum\limits_{i = 1}^m {w_i} }},
\end{equation}
where $w_{i} = \mathrm{WPNet}(f_i)$, $I_i = \mathrm{IPNet}({f_i})$.
Therefore, the quality index of $\tilde{\mathcal{P}}_A$ and $\tilde{\mathcal{P}}_B$ can be calculated through Eq. (\ref{eq:sp}) and denoted as $I(\tilde{\mathcal{P}}_A)$ and $I(\tilde{\mathcal{P}}_B)$, respectively.

Then, the ranking function $\mathscr{R}$ computes the probability that $\tilde{\mathcal{P}}_A$ is better than $\tilde{\mathcal{P}}_B$: $P_{AB} = \mathscr{R}(I(\tilde{\mathcal{P}}_A), I(\tilde{\mathcal{P}}_B))$.
Inspired by the RankNet \cite{Burgesc05,mkd17,Liuxl17,Hub19}, the overall quality indexes $I(\tilde{\mathcal{P}}_A)$ and $I(\tilde{\mathcal{P}}_B)$ are fed through a Sigmoid function to convert it into a probability output:
\begin{equation}
     P_{AB} = \frac{I(\tilde{\mathcal{P}}_A) - I(\tilde{\mathcal{P}}_B)}{1 + e^{I(\tilde{\mathcal{P}}_A) - I(\tilde{\mathcal{P}}_B)}},
\end{equation}
where $P_{AB}$ is the modeled posterior $P(I(\tilde{\mathcal{P}}_A) > I(\tilde{\mathcal{P}}_B))$.

Finally, the cross-entropy function is employed as the ranking loss $l_{r}$:
\begin{equation}
\label{eq:lab}	
    l_{r} =  - {\bar P_{AB}}\log {P_{AB}} - (1 - {\bar P_{AB}})\log (1 - {P_{AB}}),
\end{equation}
where $\bar P_{AB}$ is the desired target probability of $P_{AB}$.

\subsection{Distortion Classification Task}
\label{subsec:DCT}
%2022-TIP-Contrastive Self-Supervised Pre-Training for Video Quality Assessment.pdf D. Distortion Prediction 理论分析

The distortion classification task is introduced to distinguish among different distortion types that the inputs suffer from. 

Firstly, for each degraded point cloud $\tilde{\mathcal{P}}$, we obtain its feature vector $\tilde{f}$ via Eq. (\ref{eq:fp}).
Then, the feature vector $\tilde{f}$ is fed into a distortion classification network (DCNet), which shares the same network structure with WPNet.
Finally, the output of DCNet is passed through a classification layer with five nodes, corresponding to the five distortion types.
The cross-entropy loss is used for the distortion classification task,
\begin{equation}
    \label{eq:lc}	
    l_{c} = - \sum_{i=1}^{5} y_{i}log(\hat{y}_i),
\end{equation}
where $y_{i}$ is the one-hot encoded ground-truth vector for the $i$-th category, and $\hat{y}_i$ represents the predicted probability for the $i$-th class.

\subsection{Distortion Distribution Prediction Task}

The distortion distribution prediction task targets to predict the proportion of each type of noise in the degraded point cloud. 
Without loss of generality, we define each of the five distortion types as a weighted sum of four basic noises: Gaussian noise, uniform noise, impulse noise, and exponential noise.

Specifically, a distortion distribution prediction network (DPNet), which shares the same network structure with WPNet, is introduced.
The DPNet is followed by a soft classification head, which consists of five nodes and a Softmax function, to predict the proportion of each type of noise $\alpha$ in the degraded point clouds.
The KL divergence loss is used for the distortion distribution prediction task:
\begin{equation}
    \label{eq:ld}	
    l_{d} = \text{KL}(\alpha \parallel \hat{\alpha})
    = \sum_{k=1}^{5} \alpha_{k} \log \frac{\alpha_k}{\hat{\alpha}_k},
\end{equation}
where $\alpha$ is the true proportion of each distortion type generated during the distortion augmentation process in Section \ref{subsec:distortion_augmentation}, and $\hat{\alpha}$ represents the predicted proportions of different distortion types.

\subsection{Loss Function}

The total loss $l$ is defined as the weighted sum of the three loss functions discussed above:
\begin{equation}
\label{eq:l}	
    l = w_cL_c + w_dL_d + w_rL_{r},
\end{equation}
where $w_c$, $w_d$, and $w_r$ are the weights of these tasks, respectively.
To address the task imbalance in multi-task learning, we employ the gradient normalization (GradNorm) technique \cite{chen2018gradnorm}, which dynamically adjusts loss weights by equalizing gradient magnitudes across different tasks. 
This mechanism prioritizes complex tasks (e.g., pair-wise ranking requiring subtle distortion discrimination) through gradient-aware weight calibration, while mitigating dominance from simpler tasks. 
The self-adaptive normalization fosters balance training rates across tasks, accelerating convergence and enhancing joint feature representation learning without manual hyper-parameter tuning.

\section{Experiments} 
\label{sec:experiments}

The performance of the proposed UGD is verified from the following three aspects: two quantitative evaluations on synthetic data and one qualitative assessment on real-world data.
Firstly, the ranking accuracy is employed to quantify the UGD’s ability to distinguish between point clouds degraded by different levels of noise.% on synthetic noisy point clouds.
Then, to facilitate quantitative comparisons with traditional full-reference metrics, the UGD is used to evaluate selected typical denoising algorithms on synthetic noisy point clouds.
Finally, qualitative experiments are conducted to assess selected typical denoising algorithms on real-world noisy point clouds.

\begin{table*}[t]
\centering
\caption{Pair-wise ranking accuracy of different metrics (\%)}
\begin{tabular}{llcccccccc}
		\hline
		Type                & Method                & GN        & EN        & IN        & UN        & \multicolumn{3}{c}{MN}                        & MEAN   \\
		                    & $\alpha$              & (1,0,0,0) & (0,1,0,0) & (0,0,1,0) & (0,0,0,1) & (0.5,0.5,0,0) & (0.33,0.33,0.33,0) & (0.25,0.25,0.25,0.25) &        \\ \hline
		\multirow{7}{*}{Supervised} & $po2po_{\text{PSNR}}$ & 100.00    & 100.00    & 100.00    & 100.00    & 100.00        & 100.00        & 100.00  & 100.00\\
	  & $po2po_{\text{HD}}$   & 100.00    & 100.00     & 100.00    & 100.00    & 97.78         & 100.00        & 100.00        & 99.68  \\
	  & $po2pl_{\text{PSNR}}$ & 100.00    & 100.00    & 100.00    & 100.00    & 100.00        & 100.00        & 100.00       & 100.00 \\
	  & $po2pl_{\text{HD}}$   & 100.00    & 98.89     & 100.00    & 100.00    & 97.78         & 100.00        & 100.00        & 99.52  \\
	  & $pl2pl_{\text{PSNR}}$ & 100.00    & 100.00    & 56.67      & 100.00    & 100.00        & 58.89          & 57.78        & 81.90  \\
	  & $pl2pl_{\text{HD}}$   & 84.44     & 81.11     & 35.56     & 82.22     & 85.56          & 35.56         & 35.56         & 62.86  \\
	 & $\text{CD}$           & 100.00    & 100.00    & 100.00    & 100.00    & 100.00        & 100.00        & 100.00        & 100.00 \\
		Unsupervised        & $\text{UGD}$          & 100.00    & 98.89     & 99.44     & 100.00    & 97.78         & 98.89         & 100.00         & 99.29  \\ \hline
	\end{tabular}
\label{tab:sorting_accuracy}
\end{table*}

\subsection{Experimental Settings}

\subsubsection{Datasets}

To demonstrate the superiority of the proposed UGD, the ideal clean point cloud dataset $\mathcal{D}$, the G-PCD dataset \cite{Alexioue17}, the LiDAR-Net dataset \cite{Guo2024LiDAR-Net} and the TerraMobilita  dataset \cite{vallet2015terramobilita} are employed in the experiments.
The dataset $\mathcal{D}$ contains 150 raw point clouds, 30 of which are selected as test samples. 
The G-PCD dataset \cite{Alexioue17} consists of five raw point clouds and 25 distorted point clouds with four different levels of Gaussian noise.
The LiDAR-Net dataset~\cite{Guo2024LiDAR-Net} provides real-scanned indoor scene-level point clouds with complex structures and typical LiDAR scanning artifacts, from which one representative table scene is selected in our experiments.
The TerraMobilita real LiDAR dataset \cite{vallet2015terramobilita} is composed of 3D map scenes from LiDAR scans, from which two distinct objects are extracted as real-world noisy point cloud samples.

\subsubsection{Parameters Setting}

For the patch-wise feature learning, $m = 64$ overlapping patches with $s = 256$ points are sampled for each point cloud.
For the patch-wise prior learning, the number of Gaussian distributions $K$ is set to 8.
During the self-supervised multi-task training stage, the ideal clean point cloud dataset $\mathcal{D}$ is split such that 80\% is used for training and the remaining 20\% is reserved for testing. 
The Adam optimizer is employed for training, with a batch size of 4. 
The learning rate is initialized at $10^{-5}$ and decreased by a factor of 0.5 every 10 epochs.
The training process is conducted over a total of 100 epochs to ensure convergence and optimal performance.
Ablation studies on the parameters can be found in the supplementary material.

\subsubsection{Evaluation Criteria}
\label{subsec:metrics}

To assess the comprehensive performance of the proposed UGD, both quantitative and qualitative approaches are used in this experiment.

For the quantitative evaluation, the pair-wise ranking accuracy $\mathrm{Acc}$ is employed to evaluate the performance of UGD on the dataset $\mathcal{D}$:
\begin{equation}
\label{eq:pair-wise_accuracy}	
\mathrm{Acc} = \frac{\sum_{n=1}^{N} \mathbb{I} ((I(\mathcal{P}_A), I(\mathcal{P}_B)) \cdot y_{AB}  > 0 ) }{N},
\end{equation}
where $N$ is the total number of sampled pairs, $\mathbb{I}(x, y)$ is a function which is defined as:
\begin{equation}
\mathbb{I}(x, y) = 
\begin{cases}
    1 & \text{if } x \text{ has a higher quality than } y \\
    0 & \text{otherwise}
\end{cases},
\end{equation}
$I(\mathcal{P})$ indicates the quality score of $\mathcal{P}$ (e.g., UGD($\mathcal{P}$)),
$y_{AB}$ is the label which denotes the relative quality between $\mathcal{P}_A$ and $\mathcal{P}_B$:
\begin{equation}
y_{AB}  = 
\begin{cases}
    -1 & \text{if } \mathcal{P}_A \text{ has a higher quality than } \mathcal{P}_B  \\
    1 & \text{otherwise}
\end{cases}.
\end{equation}
It is noted that the higher the I($\mathcal{P}_A$), the higher the quality.
Besides, some commonly used traditional full-reference metrics are also included to evaluate the performance of selected denoising approaches on synthetic noisy point clouds, such as point-to-point (po2po) \cite{Rufaelm16}, point-to-plane (po2pl) \cite{tian2017geometric}, and plane-to-plane (pl2pl) \cite{Alexioue18icme}.

For the qualitative evaluation, the widely used visual inspection is employed to assess the denoising results of selected denoising methods on real-world noisy point clouds.

\subsection{Evaluation of Ranking Performance}
\label{subsec:Evaluation of Ranking Performance}

To quantitatively demonstrate the performance of the proposed UGD, we first validate its discriminative performance on synthetic noisy point clouds with different noise levels.
Specifically, each testing ground-truth clean point cloud in $\mathcal{D}$ is firstly corrupted with four basic noises (i.e., Gaussian noise, uniform noise, exponential noise, impulse noise with $\xi \in [0.05,0.1,0.15,0.2]$) and mixed noise, respectively.
Subsequently, for each type of noise, four degraded versions are paired in all possible combinations of different distortion levels.
%, resulting in $C(4, 2) = \frac{4!}{2!(4-2)!} = 6$ pairs per noise type per original point cloud.
Then, we calculate the UGD for each member of the paired samples generated above.
Finally, the pair-wise ranking accuracy $\mathrm{Acc}$ is employed to evaluate the discriminative performance of the UGD and selected full-reference metrics.

Table \ref{tab:sorting_accuracy} shows the pair-wise ranking accuracy of different metrics.
It can be seen that the majority of full-reference metrics perform well across all distortion types, except for the plane-to-plane metrics ($pl2pl_{\text{PSNR}}$ and $pl2pl_{\text{HD}}$).
As an unsupervised geometric metric, the proposed UGD performs slightly worse than the majority of full-reference metrics.
Furthermore, the UGD achieves an average ranking accuracy of 99.29\% without the need of ground-truth clean point clouds.

\begin{figure*}[!ht]
	\centering
	\includegraphics[width=1.0\textwidth]{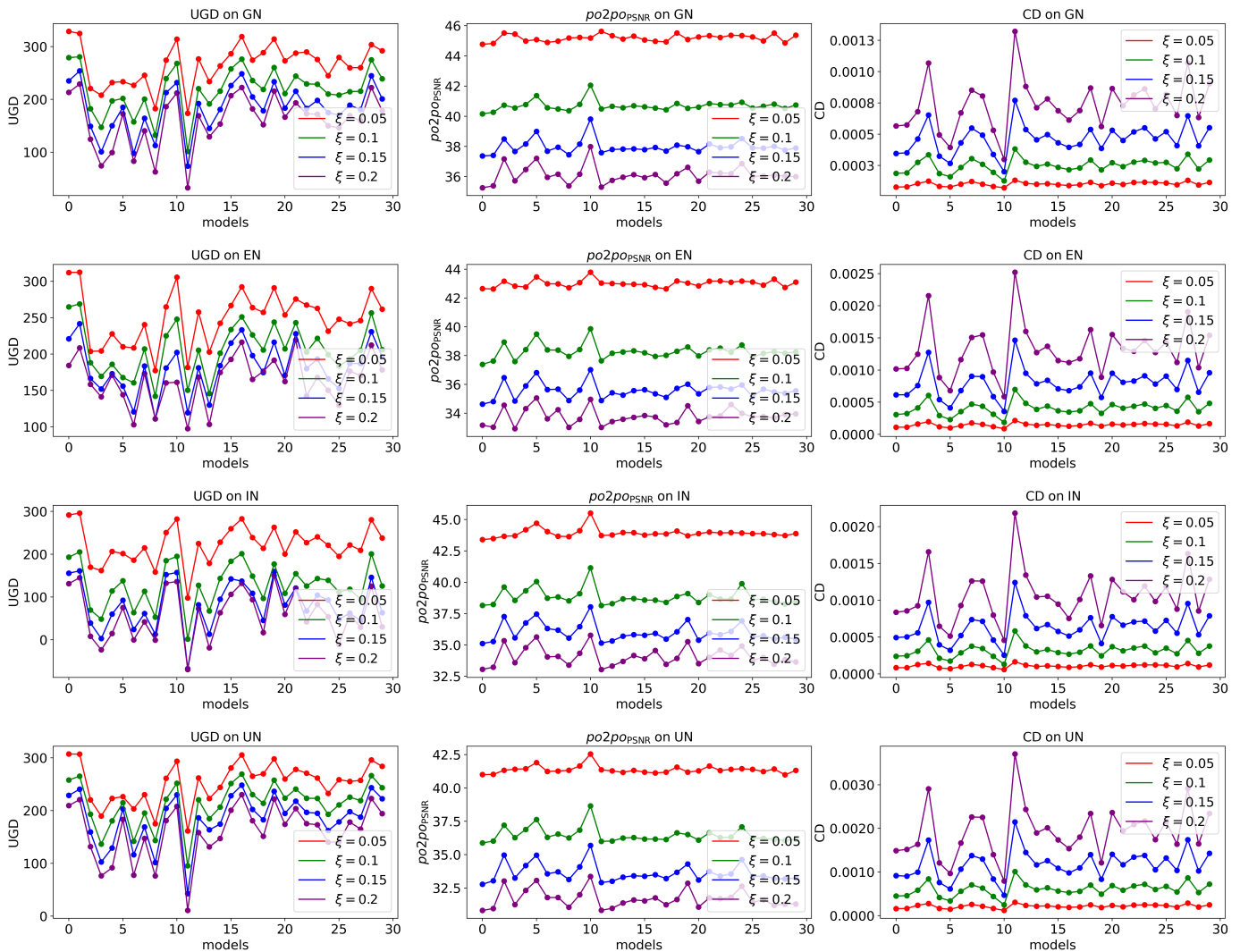}
	\caption{Visualization of results of the proposed UGD, the point-to-point (po2po) \cite{Rufaelm16} metric, and the Chamfer Distance (CD)\cite{zhang2020pointfilter}.}
	\label{fig:ArtificialNoiseVision_total_po2po}
\end{figure*}

\begin{figure}[!t]
	\centering
	\includegraphics[width=0.4\textwidth]{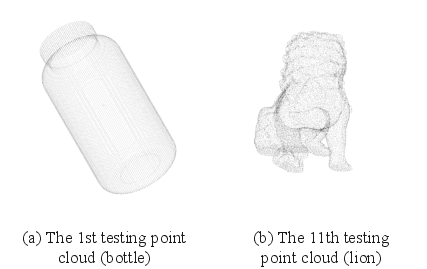}
	\caption{Visualization of structural complexity of the first and eleventh testing point clouds.}
	\label{fig:UGD_variation_analysis}
\end{figure}

To present the results more intuitively, we visualize selected results of the proposed UGD, the point-to-point (po2po) \cite{Rufaelm16} metric, and the Chamfer Distance (CD)\cite{zhang2020pointfilter} in Fig. \ref{fig:ArtificialNoiseVision_total_po2po}.
It can be observed that the proposed UGD demonstrates strong discrimination between degraded point clouds with different noise levels.
However, compared to traditional full-reference metrics, the proposed UGD shows large fluctuations in results across different testing point clouds.
Through thorough comparisons and analyses, we find that this is due to the different geometric complexity of testing point clouds.
Fig. \ref{fig:UGD_variation_analysis} visualizes the first and eleventh testing point clouds.
Obviously, from Fig. \ref{fig:ArtificialNoiseVision_total_po2po} we can find that the first point cloud (a smooth‑surfaced bottle) yields the highest UGD, whereas the eleventh point cloud (a lion) with high geometric complexity achieves the lowest UGD.
The reason behind this may be due to the fact that the proposed UGD uses the established ideal prior knowledge as a reference to compute the geometric distortion of denoised point clouds, while existing full-reference metrics all calculate the distance between the denoised point cloud and its corresponding ideal point cloud.
As a result, point clouds with smoother surfaces typically exhibit less geometric distortion, resulting in higher UGD in the unsupervised quality evaluation. 
In contrast, point clouds with complex geometric structure tend to have lower UGD due to their intricate geometric and detail distortions.
Therefore, the proposed UGD reflects not only the quality of the evaluated point cloud but also its geometric complexity.
A comprehensive breakdown of the computational complexity and runtime comparisons with supervised metrics is available in the supplementary materials.

\subsection{Evaluating Denoising Algorithms on Synthetic Noise}

In this section, we further evaluate the discriminative performance of the proposed UGD on denoising results of selected typical denoising algorithms on synthetic noisy point clouds.
We select three reference point clouds from the G-PCD dataset \cite{Alexioue17}. 
Gaussian noise with $\xi = 0.15$ is applied to the reference point clouds to generate noisy point clouds. 
The PointFilter (PF) \cite{zhang2020pointfilter}, PD-LTS \cite{Mao2024LatentSpace}, and ItertivePFN \cite{Edirimuni2023IterativePFN} are then selected to denoise the generated noisy point clouds.
Each denoised result is paired with its corresponding reference point cloud.
The denoising performance is quantitatively evaluated using both full-reference metrics (PSNR-based po2po \cite{Rufaelm16}, po2pl \cite{tian2017geometric}, pl2pl \cite{Alexioue18icme}, Chamfer Distance (CD) \cite{zhang2020pointfilter}) and the proposed UGD.

\begin{figure*}[!ht]
	\centering
	\includegraphics[width=1\textwidth]{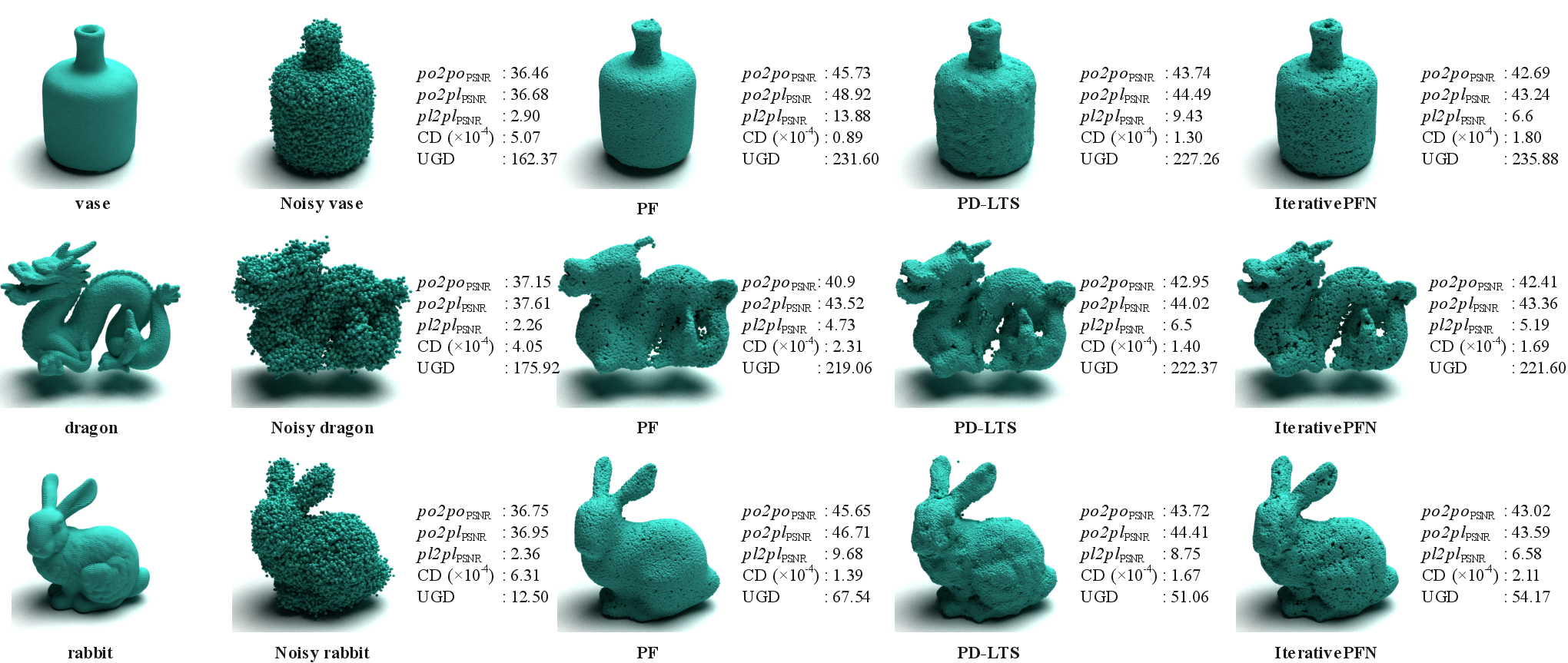}
	\caption{Visualization of denoising results of selected testing point clouds associated with evaluation metrics.}
\label{fig:DenoisingAlgorithmPerformanceEvaluationVis12}
\end{figure*}

Fig.~\ref{fig:DenoisingAlgorithmPerformanceEvaluationVis12} presents qualitative denoising results on three selected test point clouds together with the corresponding scores of different evaluation metrics. 
Overall, the three denoising methods exhibit different strengths across the cases.
The proposed UGD shows a generally similar trend to full-reference metrics in most examples, indicating good consistency with conventional objective measures.
However, an obvious inconsistency between UGD and full-reference metrics is observed on the vase model. 
Full-reference metrics tend to favor PF as the best result, whereas the visualization reveals that PF suffers from over-smoothing, which undesirably “fills” the vase mouth that should remain hollow, thus distorting the original geometry. 
In contrast, although IterativePFN achieves lower full-reference scores, it better preserves and restores the intended geometric structure (especially the critical structural details), and is therefore ranked higher by UGD.

\subsection{Evaluating Denoising Algorithms on Real-world Noise}

In this section, we assess the performance of the proposed UGD on denoising results of selected typical denoising algorithms on real-world noisy point clouds obtained from the TerraMobilita dataset~\cite{vallet2015terramobilita} (e.g., Car low and Window) and the LiDAR-Net dataset~\cite{Guo2024LiDAR-Net} (e.g., table scene).
Existing full-reference geometric metrics are no longer applicable in this scenario due to the absence of ground-truth clean reference point clouds.

\begin{figure*}[!ht]
	\centering
	\includegraphics[width=1\textwidth]{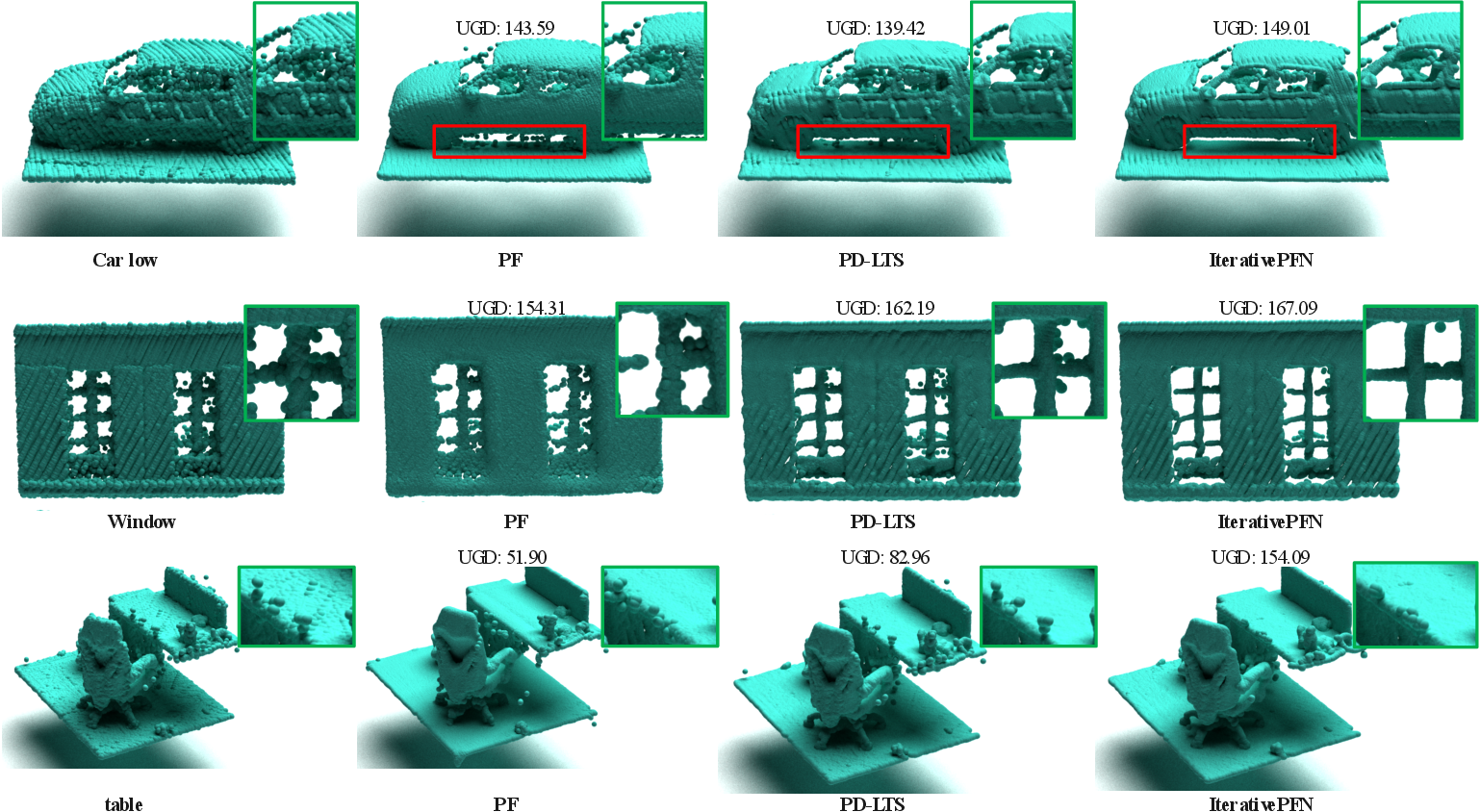}
	\caption{Visualization of denoising results of selected denoising algorithms on real-world noisy point clouds.More visualization results can be found in the supplementary material.}
	\label{fig:RealPCSubjectiveRankingVision}
\end{figure*}

Fig\ref{fig:RealPCSubjectiveRankingVision} demonstrates the effectiveness of UGD for assessing denoising quality on real-world noisy point clouds. 
% In most cases, UGD yields rankings that are consistent with the perceptual differences observed in the qualitative results.
It can be seen that the UGD is positively correlated with subjective visual quality of denoised point clouds.
For the Car low model\cite{vallet2015terramobilita}, IterativePFN provides the best overall result, effectively removing noise from both the vehicle surface and the surrounding ground.
For the Window model\cite{vallet2015terramobilita},IterativePFN still delivers the best denoising results, effectively removing the noise around the window grilles.
For the Table scene \cite{Guo2024LiDAR-Net}, which contains multiple objects, IterativePFN again delivers the best denoising performance. 
While all three methods exhibit some limitations on scene-level denoising compared with denoising a single simple model, IterativePFN remains the strongest overall and produces the cleanest results among them.

\subsection{Ablation Experiments}

\subsubsection{Impact of Weight Parameters $w_i$}

To assess the impact of weight parameters $w_i$ involved in Eq.(\ref{eq:ugd}),
we perform the experiments discussed in Section \ref{subsec:Evaluation of Ranking Performance} with and without weight parameters $w_i$, respectively.
For the latter, the weighted feature learning is disabled by directly averaging the features of each patch. 
Table \ref{tab:weight_impact} shows the ranking accuracy with and without weight parameters $w_i$.
It can be observed that incorporating weights significantly improves the performance of the UGD.

\begin{table*}[!ht]
 	\caption{Impact of weight parameters on the ranking performance of UGD  (\%)}
	\centering
	\begin{tabular}{lcccccccc}
		\hline
		                     & GN        & EN        & IN        & UN        & \multicolumn{3}{c}{MN}                        & MEAN  \\
		$\alpha$             & (1,0,0,0) & (0,1,0,0) & (0,0,1,0) & (0,0,0,1) & (0.5,0.5,0,0) & (0.33,0.33,0.33,0) & (0.25,0.25,0.25,0.25) &       \\ \hline
		UGD (with weight parameters)    & 100.00    & 98.89     & 99.44     & 100.00    & 97.78         & 98.89         & 100.00         & 99.29 \\
		UGD (without weight parameters) & 85.43     & 88.86     & 85.90     & 86.62     & 85.71         & 84.62         & 83.21         & 85.77 \\
  \hline
	\end{tabular}
	\label{tab:weight_impact}
\end{table*}

% \subsubsection{Number of Mixture Components $K$}

% To evaluate the impact of the number of mixture components $K$, firstly, the GMM is trained on the training set of $\mathcal{D}$ using different values of $K$.
% Then, we conduct the experiments discussed in Section \ref{subsec:Evaluation of Ranking Performance} with different learned GMMs.
% Finally, the pair-wise ranking accuracy is used to evaluate the discriminative performance of the UGD.
% % Subsequently, the results are compared using the pair-wise ranking accuracy on the testing set of the ideal clean point cloud  dataset $\mathcal{D}$.

% Fig. \ref{fig:k_components_accuracy} illustrates the ranking accuracy of UGD using GMMs with different numbers of mixture components. 
% It can be observed that when the number of mixture components $K$ is set to 8, the UGD achieves the highest accuracy on the testing set.
% When the number is small, fewer Gaussian models fail to fully capture the ideal prior distribution, leading to the performance degradation of UGD.

% \begin{figure}[!ht]
% 	\centering
% 	\includegraphics[width=0.5\textwidth]{k_components_accuracy.eps}
% 	\caption{Ranking performance comparison on the testing dataset with different numbers of mixture components $K$.}
% 	\label{fig:k_components_accuracy}
% \end{figure} % 

\subsubsection{Impact of Backbone Network}
To evaluate the impact of the backbone network used for patch-wise feature learning, firstly, we replace the default backbone with DGCNN \cite{dgcnn}.
Then, we conduct the experiments discussed in Section~\ref{subsec:Evaluation of Ranking Performance} under the same setting, while keeping all other components unchanged.
Finally, the pair-wise ranking accuracy is used to evaluate the discriminative performance of the UGD with different backbones.

Table~\ref{tab:backbone_ablation} reports the ranking accuracy results.
Replacing the default backbone with DGCNN leads to a decrease in MEAN ranking accuracy from $99.29\%$ to $91.67\%$, with a more noticeable degradation under mixed-noise settings (MN).
% These results suggest that the choice of the backbone affects the robustness of the learned patch-wise features.
These results indicate that a higher-performing backbone network can yield better evaluation outcomes. 
To balance algorithmic efficiency and performance, the PCT \cite{guo2021pct} is selected as the backbone network in this paper.
% Meanwhile, it is worth noting that the backbone network itself is not the focus of our contribution; in this work we use PCT as a strong default feature extractor, while the core novelty lies in the proposed pristine-prior modeling and the UGD computation framework, which can be coupled with different backbones.

\begin{table*}[!ht]
 	\caption{Backbone ablation study: pair-wise ranking accuracy of $\text{UGD}$ (\%)}
	\centering
	\begin{tabular}{lcccccccc}
		\hline
		                     & GN        & EN        & IN        & UN        & \multicolumn{3}{c}{MN}                        & MEAN  \\
		$\alpha$             & (1,0,0,0) & (0,1,0,0) & (0,0,1,0) & (0,0,0,1) & (0.5,0.5,0,0) & (0.33,0.33,0.33,0) & (0.25,0.25,0.25,0.25) &       \\ \hline
UGD (PCT)    & 100.00 & 98.89 & 99.44 & 100.00 & 97.78 & 98.89 & 100.00 & 99.29 \\
UGD (DGCNN)  & 94.44  & 97.22 & 96.67 & 98.33  & 99.44 & 80.00 & 75.56  & 91.67 \\
  \hline
	\end{tabular}
	\label{tab:backbone_ablation}
\end{table*}

\subsubsection{Multi-task Training}

%2021-NeurIPS-Efficiently Identifying Task Groupings for Multi-Task Learning
To comprehensively assess the effectiveness of the multi-task training framework, the feature encoder in the patch-wise feature learning is first trained separately on each of the three different subtasks. 
Then, the pair-wise ranking accuracy is calculated for each point cloud in the testing set of $\mathcal{D}$.

\begin{figure}[!t]
	\centering
	\includegraphics[
 trim=0 0 0 0, % 裁剪四边的白边（单位：磅）
        clip,  % 启用裁剪
        width=0.45\textwidth]{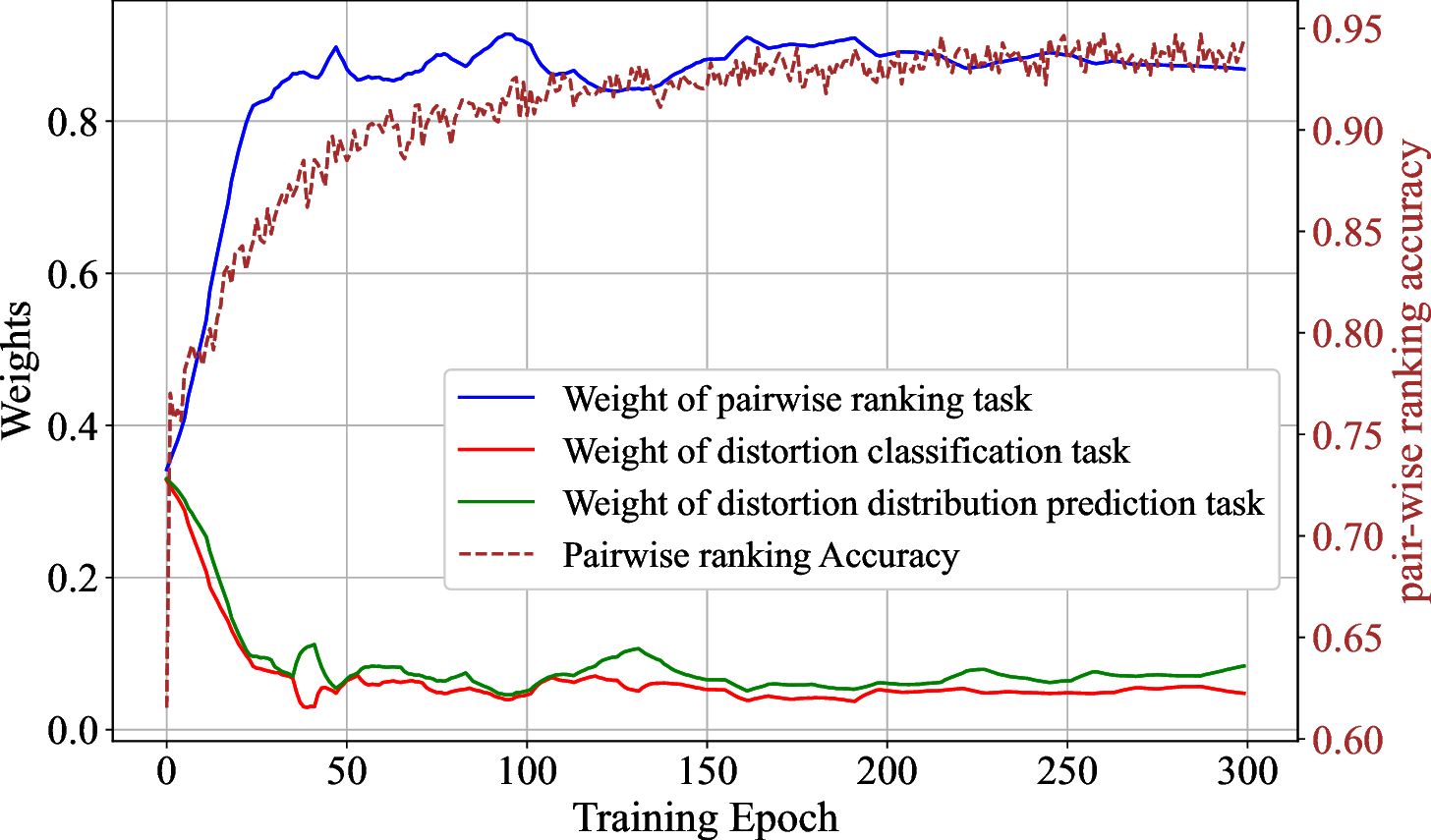}
	\caption{Visualization of the gradient normalization in multi-task training.}
	\label{fig:gradnorm_weight_en}
\end{figure}

As shown in Table \ref{tab:sppc}, the overall performance is significantly improved after applying multi-task training. 
Although the accuracy of the pair-wise ranking task is already relatively high, the multi-task training approach integrates the learning objectives of multiple subtasks, further enhancing overall performance.
Fig. \ref{fig:gradnorm_weight_en} further shows the pair-wise ranking accuracy of the UGD as a function of training time with task weights adjusted dynamically by the GradNorm.
During the initial stages of training, the GradNorm automatically assigns higher weights to the pair-wise ranking task due to its higher complexity. 
As training progresses, the weight of the pair-wise ranking task gradually increases, reflecting its dominant role in the multi-task training framework. 
This dynamic weight adjustment mechanism allows the model to focus more on the relative quality relationships in the ranking task, without getting stuck in simpler tasks such as distortion classification and distortion distribution prediction.
Therefore, the employed GradNorm strategy can achieve better overall performance and balance optimization across all tasks, particularly in the subsequent stages of training.

\begin{table*}[!ht]
	\centering
 	\caption{Comparison of the pair-wise ranking accuracy across subtasks (\%)}

\begin{tabular}{lcccccccc}
		\hline
		Subtasks                                & GN        & EN        & IN        & UN        & \multicolumn{3}{c}{MN}                        & MEAN  \\
		$\alpha$                               & (1,0,0,0) & (0,1,0,0) & (0,0,1,0) & (0,0,0,1) & (0.5,0.5,0,0) & (0.33,0.33,0.33,0) & (0.25,0.25,0.25,0.25) &       \\ \hline
		Distortion classification          & 71.45     & 68.45     & 70.02     & 72.07     & 72.07         & 74.21         & 79.62         & 72.56 \\
		Distortion distribution prediction & 80.50     & 73.48     & 80.57     & 78.81     & 77.81         & 82.69         & 83.10         & 79.56 \\
		pair-wise ranking                   & 95.19     & 94.62     & 91.98     & 92.55     & 94.60         & 94.52         & 93.64         & 93.87 \\
    		Multi-task                              & 100.00     & 98.89     & 99.44     & 100.00     & 97.78         & 98.89         & 100.00         & 99.29 \\ \hline
	\end{tabular}
	\label{tab:sppc}
\end{table*}
%TABLE%
% CONCLUSION

% \section{LIMITATION}
% While the proposed UGD demonstrates robustness across diverse scenarios, its learned“pristine prior” is inherently influenced by the distribution of the training dataset. Specifically, since high-quality clean point clouds often contain significant proportions of regular and smooth surfaces, the UGD metric may exhibit a bias towards smooth, flat, and regular structures. Consequently, when evaluating denoising methods, UGD might occasionally favor models that produce over-smoothed results at the expense of fine geometric details. Future work will focus on incorporating diverse structural complexity during prior modeling to mitigate this structural bias.

\section{Conclusion} 
\label{sec:conclusion}

In this paper, we introduce a novel unsupervised geometric distance metric for benchmarking real-world noisy point cloud denoising, computed exclusively from denoised point clouds.
The core principle underlying this approach is to develop a prior model, serving as the ground-truth, by learning from clean point clouds. 
Subsequently, the degree of degradation is quantified by measuring the variations in geometric quality between the denoised point cloud and the learned GMM priors. 
Controlled numerical experiments with synthetic noise confirm that the UGD yields comparable performance to supervised full-reference metrics.
Furthermore, we also validate the UGD on real-world noisy point clouds, and experimental results demonstrate that the proposed UGD enables unsupervised evaluation of denoising methods based exclusively on noisy data in real-world scenarios.

While the proposed UGD demonstrates robustness across diverse scenarios, its learned “pristine prior” is inherently influenced by the distribution of the training dataset. Specifically, since high-quality clean point clouds often contain significant proportions of regular and smooth surfaces, the UGD metric may exhibit a bias towards smooth, flat, and regular structures. Consequently, when evaluating denoising methods, UGD might occasionally favor models that produce over-smoothed results at the expense of fine geometric details. Future work will focus on incorporating diverse structural complexity during prior modeling to mitigate this structural bias.

% While the proposed UGD demonstrates robustness across diverse scenarios, its generalization ability relies on the diversity of the training set. Consequently, for shapes with highly specialized geometry that significantly deviate from the training set, the metric's performance might be impacted. 
% In future work, we plan to explore domain adaptation techniques to tackle this domain shift problem. 
% {\appendices
% \section*{Proof of the First Zonklar Equation}
% \section*{Proof of the Second Zonklar Equation}
% Appendix two text goes here.}

\bibliographystyle{IEEEtran}
\bibliography{main}

@ARTICLE{Wang24,
  author={Wang, Jinxi and Lu, Xuequan and Wang, Meili and Hou, Fei and He, Ying},
  journal={IEEE Transactions on Visualization and Computer Graphics}, 
  title={Learning Implicit Fields for Point Cloud Filtering}, 
  year={2025},
    volume={31},
  number={9},
   pages={5408-5420},
  doi={10.1109/TVCG.2024.3450699}}

@INPROCEEDINGS {Edirimuni2024denoise,
author = { Dasith de Silva Edirimuni and Xuequan Lu and Gang Li and Lei Wei and Antonio Robles-Kelly and Hongdong Li},
booktitle = {2024 IEEE/CVF Conference on Computer Vision and Pattern Recognition},
title = {{StraightPCF: Straight Point Cloud Filtering}},
year = {2024},
volume = {},
ISSN = {},
pages = {20721-20730}
}

@INPROCEEDINGS {Edirimuni2023IterativePFN,
author = {D. de Silva Edirimuni and X. Lu and Z. Shao and G. Li and A. Robles-Kelly and Y. He},
booktitle = { 2023 IEEE/CVF Conference on Computer Vision and Pattern Recognition},
title = {IterativePFN: True Iterative Point Cloud Filtering},
year = {2023},
volume = {},
ISSN = {},
pages = {13530-13539},
}

@INPROCEEDINGS {Luo2021Score-based,
author = { Luo, Shitong and Hu, Wei },
booktitle = { 2021 IEEE/CVF International Conference on Computer Vision },
title = { Score-Based Point Cloud Denoising },
year = {2021},
volume = {},
ISSN = {},
pages = {4563-4572},
}

@inproceedings{unknownue2022pdflow,
author = {Mao, Aihua and Du, Zihui and Wen, Yu-Hui and Xuan, Jun and Liu, Yong-Jin},
title = {PD-Flow: A Point Cloud Denoising Framework with Normalizing Flows},
year = {2022},
booktitle = { The European Conference on Computer Vision},
pages = {398-415},
}

@INPROCEEDINGS{Mao2024LatentSpace,
  author={Mao, Aihua and Yan, Biao and Ma, Zijing and He, Ying},
  booktitle={2024 IEEE/CVF Conference on Computer Vision and Pattern Recognition}, 
  title={Denoising Point Clouds in Latent Space via Graph Convolution and Invertible Neural Network}, 
  year={2024},
  volume={},
  number={},
  pages={5768-5777},
}

@INPROCEEDINGS{Guo2024LiDAR-Net,
  author={Guo, Yanwen and Li, Yuanqi and Ren, Dayong and Zhang, Xiaohong and Li, Jiawei and Pu, Liang and Ma, Changfeng and Zhan, Xiaoyu and Guo, Jie and Wei, Mingqiang and Zhang, Yan and Yu, Piaopiao and Yang, Shuangyu and Ji, Donghao and Ye, Huisheng and Sun, Hao and Liu, Yansong and Chen, Yinuo and Zhu, Jiaqi and Liu, Hongyu},
  booktitle={2024 IEEE/CVF Conference on Computer Vision and Pattern Recognition}, 
  title={LiDAR-Net: A Real-Scanned 3D Point Cloud Dataset for Indoor Scenes}, 
  year={2024},
  volume={},
  number={},
  pages={21989-21999},
 }

@article{dgcnn,
  title={Dynamic Graph CNN for Learning on Point Clouds},
  author={Wang, Yue and Sun, Yongbin and Liu, Ziwei and Sarma, Sanjay E. and Bronstein, Michael M. and Solomon, Justin M.},
  journal={ACM Transactions on Graphics},
  year={2019},
  volume = {38},
number = {5},
numpages = {12},
articleno = {146},
}

@ARTICLE{Wang2024Noise4Denoise,
  author={Wang, Weijia and Liu, Xiao and Zhou, Hailing and Wei, Lei and Deng, Zhigang and Murshed, Manzur and Lu, Xuequan},
  journal={Computational Visual Media}, 
  title={Noise4Denoise: Leveraging noise for unsupervised point cloud denoising}, 
  year={2024},
  volume={10},
  number={4},
  pages={659-669},
}

@article{Cignoni1998Metro,
  title={Metro: Measuring Error on Simplified Surfaces},
  author={ Cignoni, P.  and  Rocchini, C.  and  Scopigno, R. },
  journal={Computer Graphics Forum},
  year={1998},
  Pages ={167-174},
  volume = {17},
}

@INPROCEEDINGS{Alexioue18,
	author={Alexiou, Evangelos and Ebrahimi, Touradj},
	booktitle={2018 Picture Coding Symposium}, 
	title={Benchmarking of Objective Quality Metrics for Colorless Point Clouds}, 
	year={2018},
	volume={},
	number={},
	pages={51-55},
	doi={10.1109/PCS.2018.8456252}}

@ARTICLE{Yangq22,
	author={Yang, Qi and Ma, Zhan and Xu, Yiling and Li, Zhu and Sun, Jun},
	journal={IEEE Transactions on Pattern Analysis and Machine Intelligence}, 
	title={Inferring Point Cloud Quality via Graph Similarity}, 
	year={2022},
	volume={44},
	number={6},
	pages={3015-3029},
	doi={10.1109/TPAMI.2020.3047083}}

@ARTICLE{Zhang24TCDM,
  author={Zhang, Yujie and Yang, Qi and Zhou, Yifei and Xu, Xiaozhong and Yang, Le and Xu, Yiling},
  journal={IEEE Transactions on Visualization and Computer Graphics}, 
  title={TCDM: Transformational Complexity Based Distortion Metric for Perceptual Point Cloud Quality Assessment}, 
  year={2024},
  volume={30},
  number={10},
  pages={6707-6724},
  doi={10.1109/TVCG.2023.3338359}}

@ARTICLE{Zhang24hybrid,
  author={Zhang, Yujie and Yang, Qi and Xu, Yiling and Liu, Shan},
  journal={IEEE Transactions on Image Processing}, 
  title={Perception-Guided Quality Metric of 3D Point Clouds Using Hybrid Strategy}, 
  year={2024},
  volume={33},
  number={},
  pages={5755-5770},
  doi={10.1109/TIP.2024.3468893}}

@ARTICLE{shan24gpanet,
  author={Shan, Ziyu and Yang, Qi and Ye, Rui and Zhang, Yujie and Xu, Yiling and Xu, Xiaozhong and Liu, Shan},
  journal={IEEE Transactions on Visualization and Computer Graphics}, 
  title={GPA-Net: No-Reference Point Cloud Quality Assessment With Multi-Task Graph Convolutional Network}, 
  year={2024},
  volume={30},
  number={8},
  pages={4955-4967},
  doi={10.1109/TVCG.2023.3282802}}

@ARTICLE{zhu243dta,
  author={Zhu, Linxia and Cheng, Jun and Wang, Xu and Su, Honglei and Yang, Huan and Yuan, Hui and Korhonen, Jari},
  journal={IEEE Transactions on Multimedia}, 
  title={3DTA: No-Reference 3D Point Cloud Quality Assessment With Twin Attention}, 
  year={2024},
  volume={26},
  number={},
  pages={10489-10502},
  doi={10.1109/TMM.2024.3407698}}

@article{zyj23,
author={Y. Zhou and Z. Zhang and W. Sun and X. Min and G. Zhai},
title={ Perceptual Quality Assessment for Point Clouds: A Survey },
journal={ ZTE Communications},
year={2023},
volume={21},
number={4},
pages={3-16},
month={12},
}

@inproceedings{su2019perceptual,
title={Perceptual quality assessment of 3D point clouds},
author={Su, Honglei and Duanmu, Zhengfang and Liu, Wentao and Liu, Qi and Wang, Zhou},
booktitle={2019 IEEE International Conference on Image Processing},
pages={3182--3186},
year={2019}
}

@article{dempster1977maximum,
  title={Maximum likelihood from incomplete data via the EM algorithm},
  author={Dempster, Arthur P. and Laird, Nan M. and Rubin, Donald B.},
  journal={Journal of the Royal Statistical Society. Series B (Methodological)},
  volume={39},
  number={1},
  pages={1--22},
  year={1977},
  publisher={Wiley Online Library}
}

@article{viola2020reduced,
  title={A reduced reference metric for visual quality evaluation of point cloud contents},
  author={Viola, Irene and Cesar, Pablo},
  journal={IEEE Signal Processing Letters},
  volume={27},
  pages={1660--1664},
  year={2020},
  publisher={IEEE}
}

@article{zhou2023reduced,
  title={Reduced-reference quality assessment of point clouds via content-oriented saliency projection},
  author={Zhou, Wei and Yue, Guanghui and Zhang, Ruizeng and Qin, Yipeng and Liu, Hantao},
  journal={IEEE Signal Processing Letters},
  volume={30},
  pages={354--358},
  year={2023},
  publisher={IEEE}
}

@ARTICLE{wuxj21,
	author={Wu, Xinju and Zhang, Yun and Fan, Chunling and Hou, Junhui and Kwong, Sam},
	journal={IEEE Transactions on Circuits and Systems for Video Technology}, 
	title={Subjective Quality Database and Objective Study of Compressed Point Clouds With 6DoF Head-Mounted Display}, 
	year={2021},
	volume={31},
	number={12},
	pages={4630-4644},
	doi={10.1109/TCSVT.2021.3101484}}

@ARTICLE{Yangq21,
	author={Yang, Qi and Chen, Hao and Ma, Zhan and Xu, Yiling and Tang, Rongjun and Sun, Jun},
	journal={IEEE Transactions on Multimedia}, 
	title={Predicting the Perceptual Quality of Point Cloud: A 3D-to-2D Projection-Based Exploration}, 
	year={2021},
	volume={23},
	number={},
	pages={3877-3891},
	doi={10.1109/TMM.2020.3033117}}

@inproceedings{chen2018gradnorm,
  title={Gradnorm: Gradient normalization for adaptive loss balancing in deep multitask networks},
  author={Chen, Zhao and Badrinarayanan, Vijay and Lee, Chen-Yu and Rabinovich, Andrew},
  booktitle={International Conference on Machine Learning},
  pages={794-803},
  year={2018}
}

@article{liu2021pqa,
  title={PQA-Net: Deep no reference point cloud quality assessment via multi-view projection},
  author={Liu, Qi and Yuan, Hui and Su, Honglei and Liu, Hao and Wang, Yu and Yang, Huan and Hou, Junhui},
  journal={IEEE Transactions on Circuits and Systems for Video Technology},
  volume={31},
  number={12},
  pages={4645--4660},
  year={2021},
  publisher={IEEE}
}

@inproceedings{yang2022no,
  title={No-reference point cloud quality assessment via domain adaptation},
  author={Yang, Qi and Liu, Yipeng and Chen, Siheng and Xu, Yiling and Sun, Jun},
  booktitle={Proceedings of the IEEE/CVF Conference on Computer Vision and Pattern Recognition},
  pages={21179--21188},
  year={2022}
}

@article{zhang24gms,
author = {Zhang, Zicheng and Sun, Wei and Wu, Haoning and Zhou, Yingjie and Li, Chunyi and Chen, Zijian and Min, Xiongkuo and Zhai, Guangtao and Lin, Weisi},
title = {GMS-3DQA: Projection-Based Grid Mini-patch Sampling for 3D Model Quality Assessment},
year = {2024},
volume = {20},
number = {6},
journal = {ACM Transactions on Multimedia Computing Communications and Applications},
articleno = {178},
numpages = {19},
}

@article{lavoue2010comparison,
  title={A comparison of perceptually-based metrics for objective evaluation of geometry processing},
  author={Lavoué, Guillaume and Corsini, Massimiliano},
  journal={IEEE Transactions on Multimedia},
  volume={12},
  number={7},
  pages={636--649},
  year={2010},
  publisher={IEEE}
}

@inproceedings{javaheri2020generalized,
  title={A generalized Hausdorff distance based quality metric for point cloud geometry},
  author={Javaheri, Alireza and Brites, Catarina and Pereira, Fernando and Ascenso, Jo{\~a}o},
  booktitle={2020 Twelfth International Conference on Quality of Multimedia Experience},
  pages={1--6},
  year={2020}
}

@inproceedings{javaheri2017subjective,
  title={Subjective and objective quality evaluation of 3D point cloud denoising algorithms},
  author={Javaheri, Alireza and Brites, Catarina and Pereira, Fernando and Ascenso, Jo{\~a}o},
  booktitle={2017 IEEE International Conference on Multimedia \& Expo Workshops},
  pages={1--6},
  year={2017}
}

@article{garg2019survey,
  title={A survey of denoising techniques for multi-parametric prostate MRI},
  author={Garg, Gaurav and Juneja, Mamta},
  journal={Multimedia Tools and Applications},
  volume={78},
  pages={12689--12722},
  year={2019}
}

@inproceedings{tian2017geometric,
  title={Geometric distortion metrics for point cloud compression},
  author={Tian, Dong and Ochimizu, Hideaki and Feng, Chen and Cohen, Robert and Vetro, Anthony},
  booktitle={2017 IEEE International Conference on Image Processing},
  pages={3460--3464},
  year={2017}
}

@article{liu2021reduced,
  title={Reduced reference perceptual quality model with application to rate control for video-based point cloud compression},
  author={Liu, Qi and Yuan, Hui and Hamzaoui, Raouf and Su, Honglei and Hou, Junhui and Yang, Huan},
  journal={IEEE Transactions on Image Processing},
  volume={30},
  pages={6623--6636},
  year={2021}
}

@article{rakotosaona2020pointcleannet,
  title={Pointcleannet: Learning to denoise and remove outliers from dense point clouds},
  author={Rakotosaona, Marie-Julie and La Barbera, Vittorio and Guerrero, Paul and Mitra, Niloy J and Ovsjanikov, Maks},
  journal={Computer Graphics Forum},
  volume={39},
  number={1},
  pages={185--203},
  year={2020}
}

@article{zhang2020pointfilter,
  title={Pointfilter: Point cloud filtering via encoder-decoder modeling},
  author={Zhang, Dongbo and Lu, Xuequan and Qin, Hong and He, Ying},
  journal={IEEE Transactions on Visualization and Computer Graphics},
  volume={27},
  number={3},
  pages={2015--2027},
  year={2020}
}

@inproceedings{hermosilla2019total,
  title={Total Denoising: Unsupervised Learning of 3D Point Cloud Cleaning},
  author={Hermosilla, Pedro and Ritschel, Tobias and Ropinski, Timo},
  booktitle={Proceedings of the IEEE/CVF International Conference on Computer Vision},
  pages={52--60},
  year={2019}
}

@article{Huang2023MODNet,
  title={MODNet: Multi-offset Point Cloud Denoising Network Customized for Multi-scale Patches},
  author = {Anyi Huang and Qian Xie and Zhoutao Wang and Dening Lu and Mingqiang Wei and Jun Wang},
  journal={Computer Graphics Forum},
  volume={41},
    Isssue={7},
  pages={109-119},
  year={2023}
}

@inproceedings{TurkG94, 
	author = {Turk, Greg and Levoy, Marc}, 
	title = {Zippered Polygon Meshes from Range Images}, 
	year = {1994}, 
	booktitle = {Proceedings of the 21st Annual Conference on Computer Graphics and Interactive Techniques}, 
	pages = {311-318}, 
	numpages = {8},
 }

@INPROCEEDINGS{Wuzr15,

	author={Zhirong Wu and Song, Shuran and Khosla, Aditya and Fisher Yu and Linguang Zhang and Xiaoou Tang and Xiao, Jianxiong},
	booktitle={2015 IEEE Conference on Computer Vision and Pattern Recognition}, 
	title={3D ShapeNets: A deep representation for volumetric shapes}, 
	year={2015},
	volume={},
	number={},
	pages={1912-1920}}

@inproceedings{qi2017pointnet++,
	title     = {PointNet++: Deep hierarchical feature learning on point sets in a metric space},
	booktitle = {Proceedings of the 31th International Conference on Neural Information Processing Systems},
	author={Qi, Charles Ruizhongtai and Yi, Li and Su, Hao and Guibas, Leonidas J},
	pages={5099-5108},
	year={2017}
}

@ARTICLE{liuq23,
	author={Liu, Qi and Su, Honglei and Duanmu, Zhengfang and Liu, Wentao and Wang, Zhou},
	journal={IEEE Transactions on Visualization and Computer Graphics}, 
	title={Perceptual Quality Assessment of Colored 3D Point Clouds}, 
	year={2023},
  volume={29},
  number={8},
  pages={3642-3655}}

@article{Liu2023,
author = {Liu, Yipeng and Yang, Qi and Xu, Yiling and Yang, Le},
title = {Point Cloud Quality Assessment: Dataset Construction and Learning-based No-reference Metric},
year = {2023},
publisher = {Association for Computing Machinery},
address = {New York, NY, USA},
volume = {19},
number = {2s},
journal = {ACM Transactions on Multimedia Computing Communications and Applications},
articleno = {80},
numpages = {26}
}

@inproceedings{Burgesc05,
	title={Learning to rank using gradient descent},
	author={ Burges, Chris  and  Shaked, Tal  and  Renshaw, Erin  and  Lazier, Ari  and  Deeds, Matt  and  Hamilton, Nicole  and  Hullender, Greg },
	booktitle={Proceedings of the 22nd international conference on Machine learning},
	pages={89-96},
	year={2005},
	
}

@ARTICLE{mkd17,
	author={Ma, Kede and Liu, Wentao and Liu, Tongliang and Wang, Zhou and Tao, Dacheng},
	journal={IEEE Transactions on Image Processing}, 
	title={dipIQ: Blind Image Quality Assessment by Learning-to-Rank Discriminable Image Pairs}, 
	year={2017},
	volume={26},
	number={8},
	pages={3951-3964}}

@inproceedings{Liuxl17,
	title={RankIQA: Learning from rankings for no-reference image quality assessment}, 
	author={Liu, Xialei and Van De Weijer, Joost and Bagdanov, Andrew D},
	booktitle={Proceedings of the IEEE International Conference on Computer Vision},
	pages={1040-1049},
	year={2017},

}

@ARTICLE{Hub19,
	author={Hu, Bo and Li, Leida and Liu, Hantao and Lin, Weisi and Qian, Jiansheng},
	journal={IEEE Transactions on Multimedia}, 
	title={Pairwise-Comparison-Based Rank Learning for Benchmarking Image Restoration Algorithms}, 
	year={2019},
	volume={21},
	number={8},
	pages={2042-2056}}

@article{liu2022no,
  title={No-reference bitstream-layer model for perceptual quality assessment of V-PCC encoded point clouds},
  author={Liu, Qi and Su, Honglei and Chen, Tianxin and Yuan, Hui and Hamzaoui, Raouf},
  journal={IEEE Transactions on Multimedia},
  volume={25},
  pages={4533--4546},
  year={2022},
  publisher={IEEE}
}

@INPROCEEDINGS {shan2024contrastive,
author = { Shan, Ziyu and Zhang, Yujie and Yang, Qi and Yang, Haichen and Xu, Yiling and Hwang, Jenq-Neng and Xu, Xiaozhong and Liu, Shan },
booktitle = { 2024 IEEE/CVF Conference on Computer Vision and Pattern Recognition},
title = {{ Contrastive Pre-Training with Multi-View Fusion for No-Reference Point Cloud Quality Assessment }},
year = {2024},
volume = {},
ISSN = {},
pages = {25942-25951}}

@INPROCEEDINGS{marcos2023evaluating,
  title={Evaluating unsupervised denoising requires unsupervised metrics},
  author={Marcos-Morales, Adria and Leibovich, Matan and Mohan, Sreyas and Vincent, Joshua Lawrence and Haluai, Piyush and Tan, Mai and Crozier, Peter and Fernandez-Granda, Carlos},
year = {2023},
booktitle={Proceedings of the 40th International Conference on Machine Learning},
pages={23937--23957},
location = {Honolulu, Hawaii, USA}
}

@article{guo2021pct,
  title={Pct: Point cloud transformer},
  author={Guo, Meng-Hao and Cai, Jun-Xiong and Liu, Zheng-Ning and Mu, Tai-Jiang and Martin, Ralph R and Hu, Shi-Min},
  journal={Computational Visual Media},
  volume={7},
  pages={187--199},
  year={2021},
  publisher={Springer}
}

@INPROCEEDINGS{Alexioue17,

	author={Alexiou, Evangelos and Upenik, Evgeniy and Ebrahimi, Touradj},
	booktitle={2017 IEEE 19th International Workshop on Multimedia Signal Processing (MMSP)}, 
	title={Towards subjective quality assessment of point cloud imaging in augmented reality}, 
	year={2017},
	volume={},
	number={},
	pages={1-6}
}

@article{vallet2015terramobilita,
  title={TerraMobilita/iQmulus urban point cloud analysis benchmark},
  author={Vallet, Bruno and Br{\'e}dif, Mathieu and Serna, Andr{\'e}s and Marcotegui, Beatriz and Paparoditis, Nicolas},
  journal={Computers \& Graphics},
  volume={49},
  pages={126--133},
  year={2015},
  publisher={Elsevier}
}

@article{wei2024pathnet,
  title={PathNet: Path-selective point cloud denoising},
  author={Z. Wei and H. Chen and L. Nan and J. Wang and J. Qin and M. Wei},
  journal={IEEE Transactions on Pattern Analysis and Machine Intelligence},
  volume={46},
  number={6},
  pages={4426-4442},year={2024},
  publisher={IEEE}
}

@article{feng2024forest,
  title={Forest point cloud denoising method based on curvature and density},
  author={Feng, Qian and Zhu, Tingting and Ni, Chao},
  journal={International Journal of Remote Sensing},
  volume={45},
  number={22},
  pages={8105--8122},
  year={2024},
  publisher={Taylor \& Francis}
}

@article{su2025cad,
title = {SITF: A Self-Supervised Iterative Training Framework for Point Cloud Denoising},
journal = {Computer-Aided Design},
volume = {179},
pages = {103812},
year = {2025},
issn = {0010-4485},
doi = {https://doi.org/10.1016/j.cad.2024.103812},
author = {Zhiyong Su and Changchang Wang and Kun Jiang and Kai Jiang and Weiqing Li}
}

@article{Zhoul22,
title = {Point cloud denoising review: from classical to deep learning-based approaches},
journal = {Graphical Models},
volume = {121},
pages = {101140},
year = {2022},
issn = {1524-0703},
author = {Lang Zhou and Guoxing Sun and Yong Li and Weiqing Li and Zhiyong Su}
}

@article{alexa-2003-MLS,
  author={Alexa, M. and Behr, J. and Cohen-Or, D. and Fleishman, S. and Levin, D. and Silva, C.T.},
  journal={IEEE Transactions on Visualization and Computer Graphics}, 
  title={Computing and rendering point set surfaces}, 
  year={2003},
  volume={9},
  number={1},
  pages={3-15}}

@article{fleishman-2005-RMLS,
  title={Robust moving least-squares fitting with sharp features},
  author={Fleishman, Shachar and Cohen-Or, Daniel and Silva, Cl{\'a}udio T},
  journal={ACM Transactions on Graphics},
  volume={24},
  number={3},
  pages={544--552},
  year={2005},
  publisher={ACM New York, NY, USA}
}

@article{oztireli-2009-RIMLS,
  title={Feature preserving point set surfaces based on non-linear kernel regression},
  author={Oztireli, Cengiz and Guennebaud, Gael and Gross, Markus},
  journal={Computer Graphics Forum},
  volume={28},
  number={2},
  pages={493--501},
  year={2009}
}

@article{lipman-2007-LOP,
  title={Parameterization-free projection for geometry reconstruction},
  author={Lipman, Yaron and Cohen-Or, Daniel and Levin, David and Tal-Ezer, Hillel},
  journal={ACM Transactions on Graphics},
  volume={26},
  number={3},
  pages={22},
  year={2007},
 
}

@article{huang-2009-WLOP,
  title={Consolidation of unorganized point clouds for surface reconstruction},
  author={Huang, Hui and Li, Dan and Zhang, Hao and Ascher, Uri and Cohen-Or, Daniel},
  journal={ACM Transactions on Graphics},
  volume={28},
  number={5},
  pages={1--7},
  year={2009},
  publisher={ACM New York, NY, USA}
}

@article{preiner-2014-CLOP,
  title={Continuous projection for fast L1 reconstruction},
  author={Preiner, Reinhold and Mattausch, Oliver and Arikan, Murat and Pajarola, Renato and Wimmer, Michael},
  journal={ACM Transactions on Graphics},
  articleno = {47},
pages = {1-13},
volume = {33},
  year={2014}
}

@article{guerrero-2018-pcpnet,
  title={Pcpnet learning local shape properties from raw point clouds},
  author={Guerrero, Paul and Kleiman, Yanir and Ovsjanikov, Maks and Mitra, Niloy J},
  journal={Computer Graphics Forum},
  volume={37},
  number={2},
  pages={75-85},
  year={2018}
}

@article{wei-2021-geodualcnn,
  author={Wei, Mingqiang and Chen, Honghua and Zhang, Yingkui and Xie, Haoran and Guo, Yanwen and Wang, Jun},
  journal={IEEE Transactions on Visualization and Computer Graphics}, 
  title={GeoDualCNN: Geometry-Supporting Dual Convolutional Neural Network for Noisy Point Clouds}, 
  year={2023},
  volume={29},
  number={2},
  pages={1357-1370},
  doi={10.1109/TVCG.2021.3113463}}

@inproceedings{luo-2020-DMRDenoise,
  title={Differentiable manifold reconstruction for point cloud denoising},
  author={Luo, Shitong and Hu, Wei},
  booktitle={Proceedings of the 28th ACM International Conference on Multimedia },
  pages={1330--1338},
  year={2020}
}

@ARTICLE{edirimuni-2024-contrastive,
    author={D. de Silva Edirimuni and X. Lu and G. Li and A. Robles-Kelly},
  journal={IEEE Transactions on Visualization and Computer Graphics}, 
  title={Contrastive Learning for Joint Normal Estimation and Point Cloud Filtering}, 
  year={2024},
  volume={30},
  number={8},
  pages={4527-4541}}

@article{Rufaelm16,
	title={Evaluation criteria for PCC (Point Cloud Compression)},
	author={Rufael Mekuria and Zhu Li and Christian Tulvan and Philip A. Chou},
	journal = {ISO/IEC MPEG N16332},
	year={2016}
}

@inproceedings{Alexioue18icme,
	title={Point Cloud Quality Assessment Metric Based on Angular Similarity},
	author={Evangelos Alexiou and Touradj Ebrahimi},
	booktitle={2018 IEEE International Conference on Multimedia and Expo},
	pages={1-6},
	year={2018}
}
\begin{IEEEbiography}[{\includegraphics[width=1in,height=1.25in,clip,keepaspectratio]
{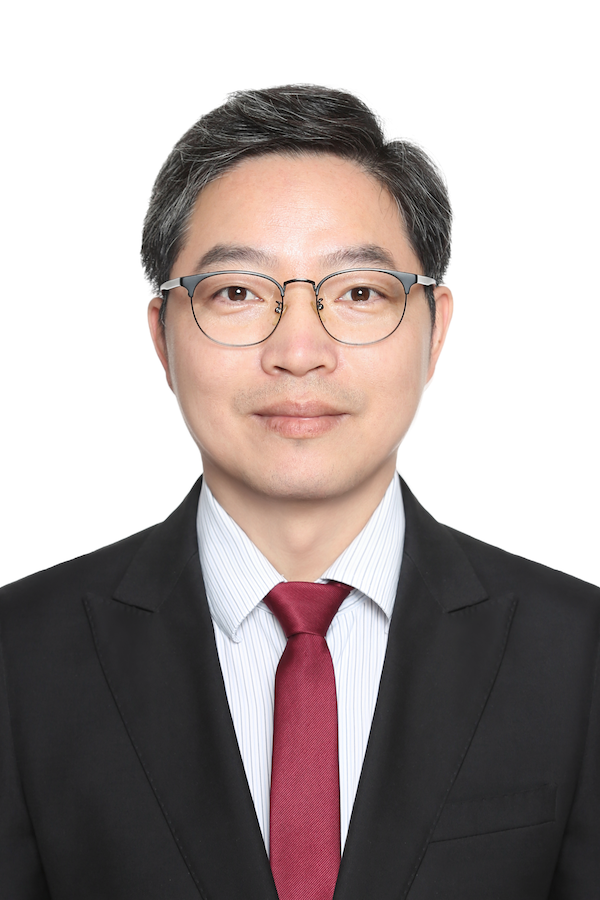}}]{Zhiyong Su}
is currently an associate professor at the School of Automation, Nanjing University of Science and Technology, China. He received the B.S. and M.S. degrees from the School of Computer Science and Technology, Nanjing University of Science and Technology in 2004 and 2006, respectively, and received the Ph.D. from the Institute of Computing Technology, Chinese Academy of Sciences in 2009. His current interests include computer graphics, computer vision, augmented reality, and machine learning.
\end{IEEEbiography}

\begin{IEEEbiography}[{\includegraphics[width=1in,height=1.25in,clip,keepaspectratio]{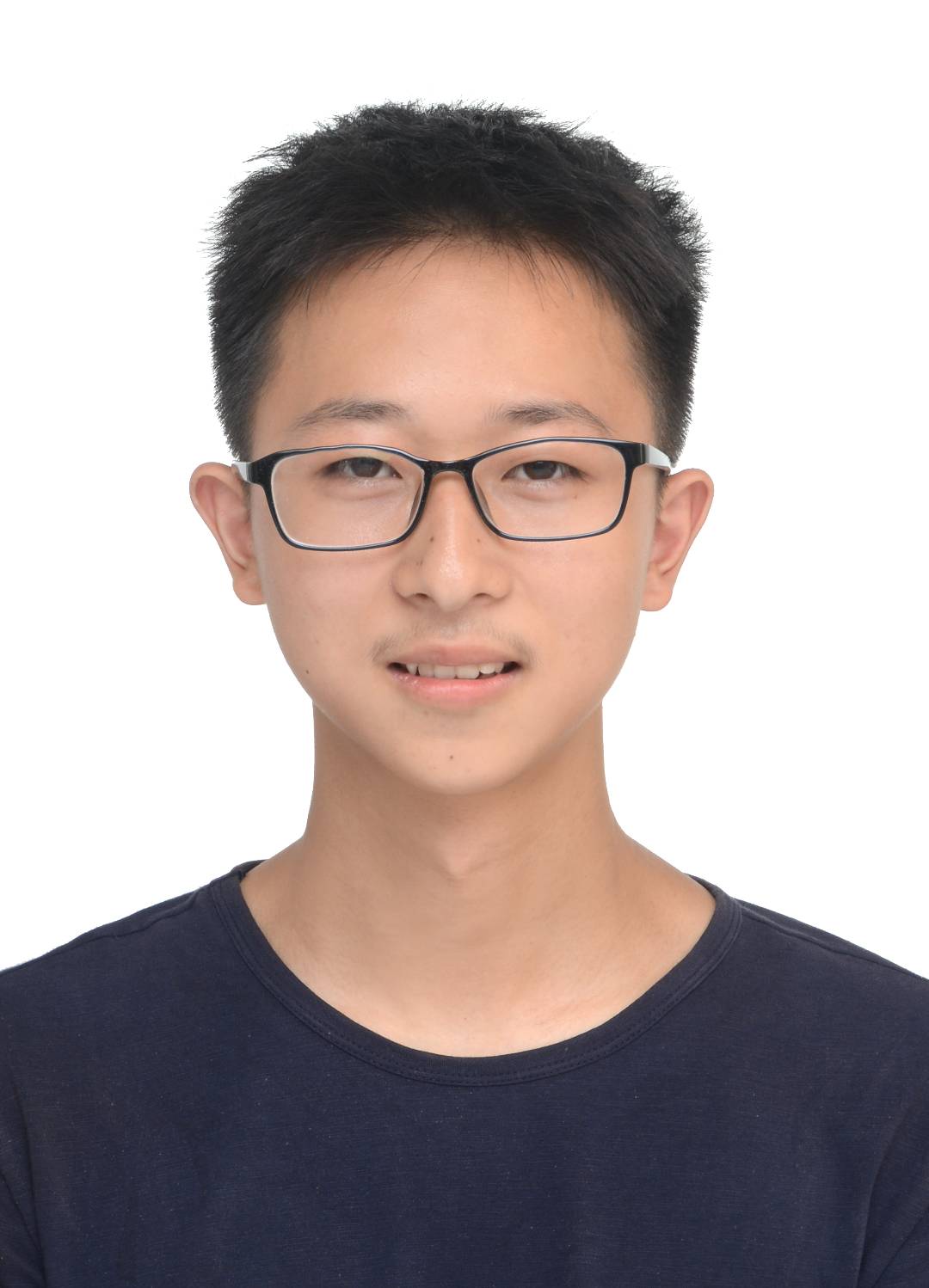}}]{Jincan Wu}
received the B.S. degree from Nanjing University of Science and Technology, Jiangsu, China in 2023 and now studies at Nanjing University of Science and Technology. His research interests include mesh quality assessment and 3DGS quality assessment.
\end{IEEEbiography}

\begin{IEEEbiography}[{\includegraphics[width=1in,height=1.25in,clip,keepaspectratio]{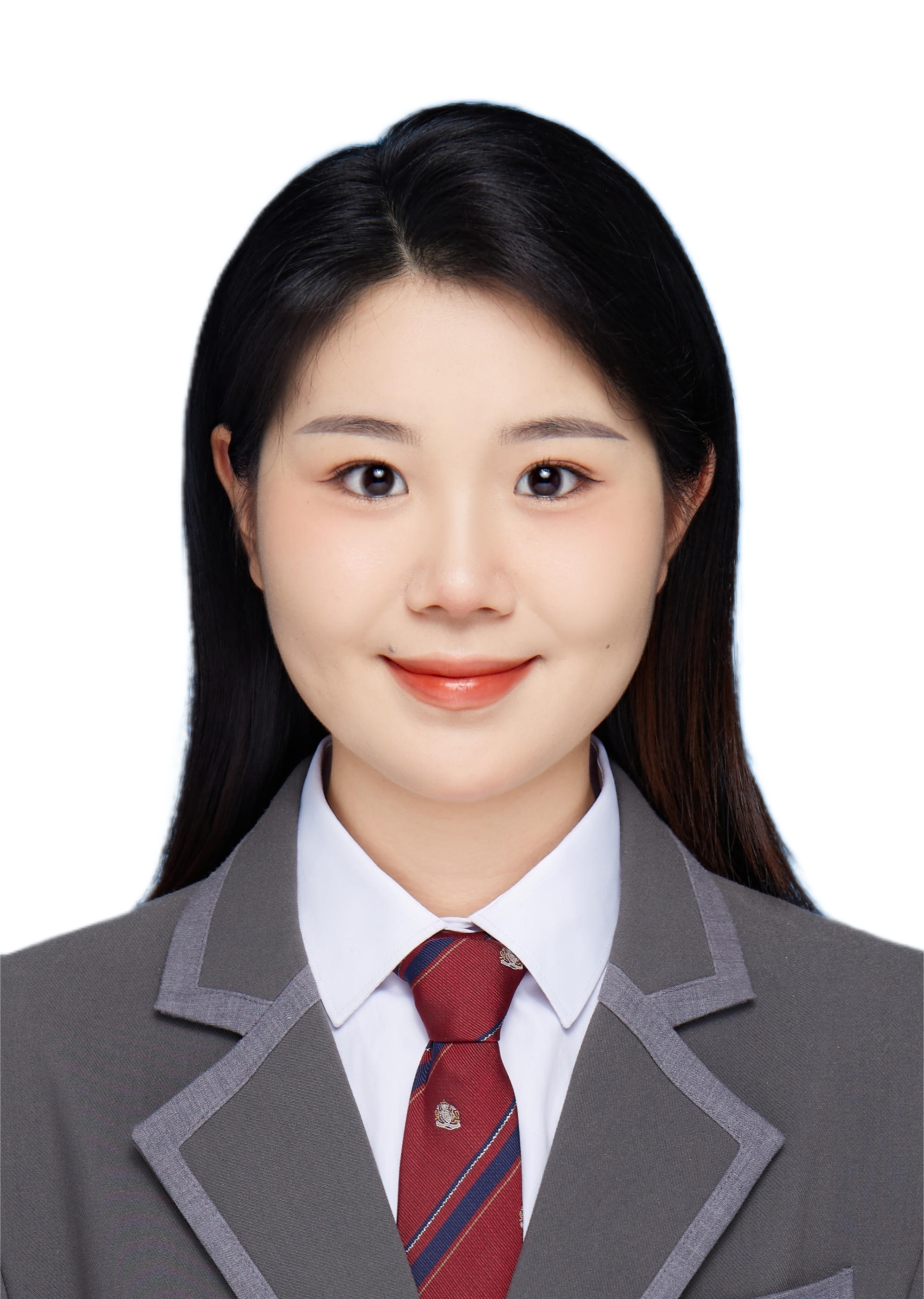}}]{Yanke Li}
received the B.S. degree from Hefei University of Technology, Anhui, China in 2024 and now studies at Nanjing University of Science and Technology. Her research interests include 3DGS quality assessment and denoising.
\end{IEEEbiography}

\begin{IEEEbiography}[{\includegraphics[width=1in,height=1.25in,clip,keepaspectratio]{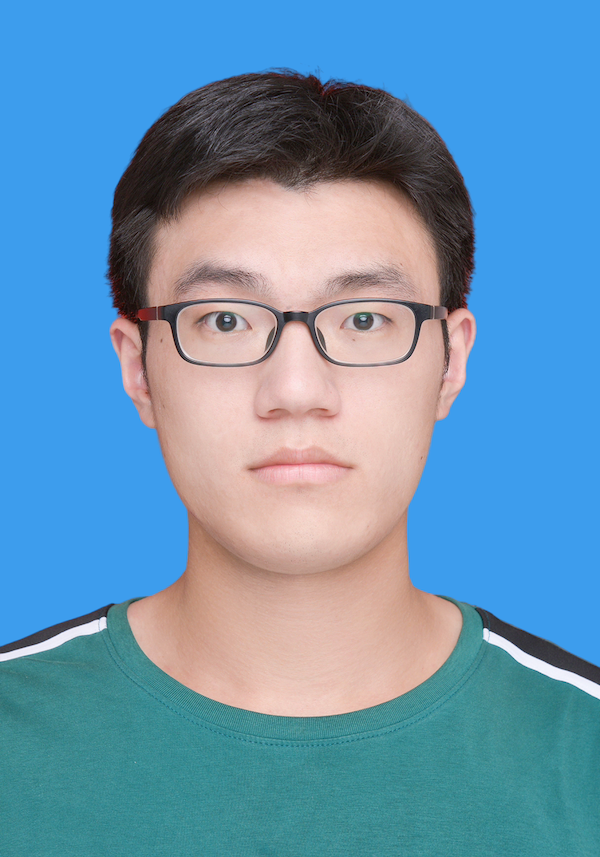}}]{Yonghui Liu}
received the B.S. degree from China Jiliang University, Hangzhou, China in 2024 and now studies at Nanjing University of Science and Technology. His research interests include infrared image simulation and image quality assessment.
\end{IEEEbiography}

\begin{IEEEbiography}[{\includegraphics[width=1in,height=1.25in,clip,keepaspectratio]{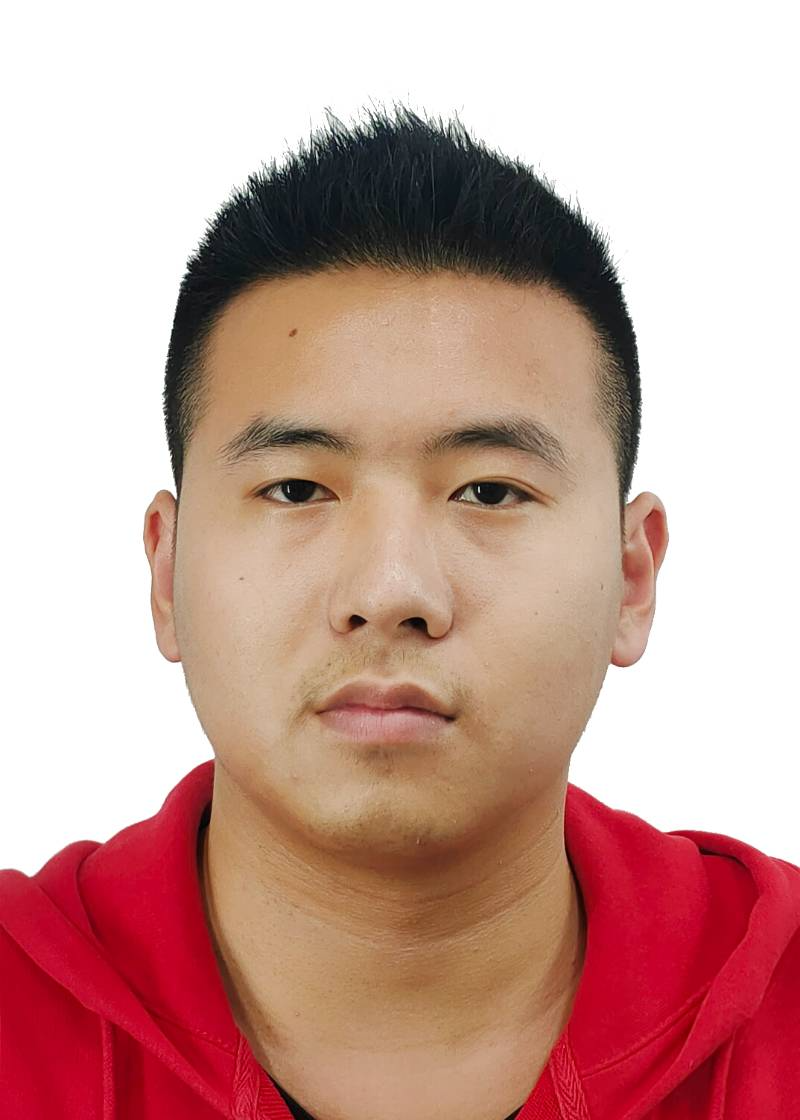}}]{Zheng Li}
received the B.S. degree from Xuzhou University of Technology, Jiangsu, China in 2022 and now studies at Nanjing University of Science and Technology. His research interests include quality assessment of point clouds.
\end{IEEEbiography}

\begin{IEEEbiography}[{\includegraphics[width=1in,height=1.25in,clip,keepaspectratio]
{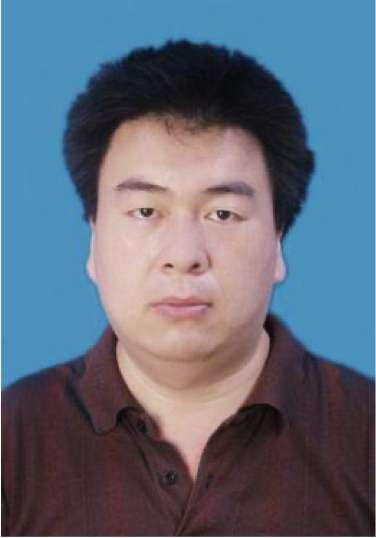}}]{Weiqing Li}
is currently an associate professor at the School of Computer Science and Engineering, Nanjing University of Science and Technology, China. He received the B.S. and Ph.D. degrees from the School of Computer Sciences and Engineering, Nanjing University of Science and Technology in 1997 and 2007, respectively. His current interests include computer graphics and virtual reality.
\end{IEEEbiography}

%\vfill

\end{document}